\documentclass[11pt]{article}

\usepackage[T1]{fontenc}
\usepackage[utf8]{inputenc}
\usepackage{tgpagella}
\usepackage{mathpazo}
\usepackage{inconsolata}
\usepackage{amsmath}
\usepackage{amssymb}
\usepackage[nopatch=footnote]{microtype}
\usepackage{booktabs}
\usepackage{graphicx}
\usepackage[percent]{overpic}
\usepackage{caption}
\usepackage{tabularx}
\usepackage{array}
\usepackage{multirow}
\usepackage{makecell}
\usepackage[numbers]{natbib}
\usepackage{ovisocr2_report}

\newcolumntype{Y}{>{\raggedright\arraybackslash}X}

\newcommand{\ours}{OvisOCR2}
\newcommand{\up}{\(\uparrow\)}
\newcommand{\down}{\(\downarrow\)}
\newcommand{\bestcell}[1]{\textbf{#1}}
\newcommand{\secondcell}[1]{\underline{#1}}
\title{OvisOCR2 Technical Report}
\author{
  ATH-MaaS, Alibaba Group\\[0.4em]
  \small Full author list is provided at the end of this report.
}
\date{}

\begin{document}

\maketitle

\begin{abstract}
We introduce OvisOCR2, a 0.8B document parsing model. OvisOCR2 is designed as an end-to-end parser: given a document page image, it generates a Markdown representation in natural reading order, covering text, formulas, tables, and visual regions. We build a data engine that combines filtered real-document annotations with synthetic pages whose rendered images and Markdown targets are derived from the same HTML source. The training recipe includes supervised fine-tuning, reinforcement learning on a 4B branch with a multi-component reward design, on-policy distillation into the 0.8B model, and model fusion. On OmniDocBench v1.6, OvisOCR2 achieves a state-of-the-art overall score of 96.58, placing an end-to-end model at the top of this leaderboard previously dominated by pipeline methods and highlighting the potential of end-to-end document parsing. On PureDocBench, OvisOCR2 also achieves the highest Avg3 score of 75.06. Beyond these two public benchmarks, we evaluate OvisOCR2 on an in-house benchmark designed to cover a broader set of long-tail and challenging scenarios. OvisOCR2 obtains the best overall performance among the compared methods, providing further evidence of its generalization and robustness. OvisOCR2 is available at \url{https://huggingface.co/ATH-MaaS/OvisOCR2}.

\end{abstract}

\begin{figure}[!htbp]
\centering
\includegraphics[width=0.88\linewidth]{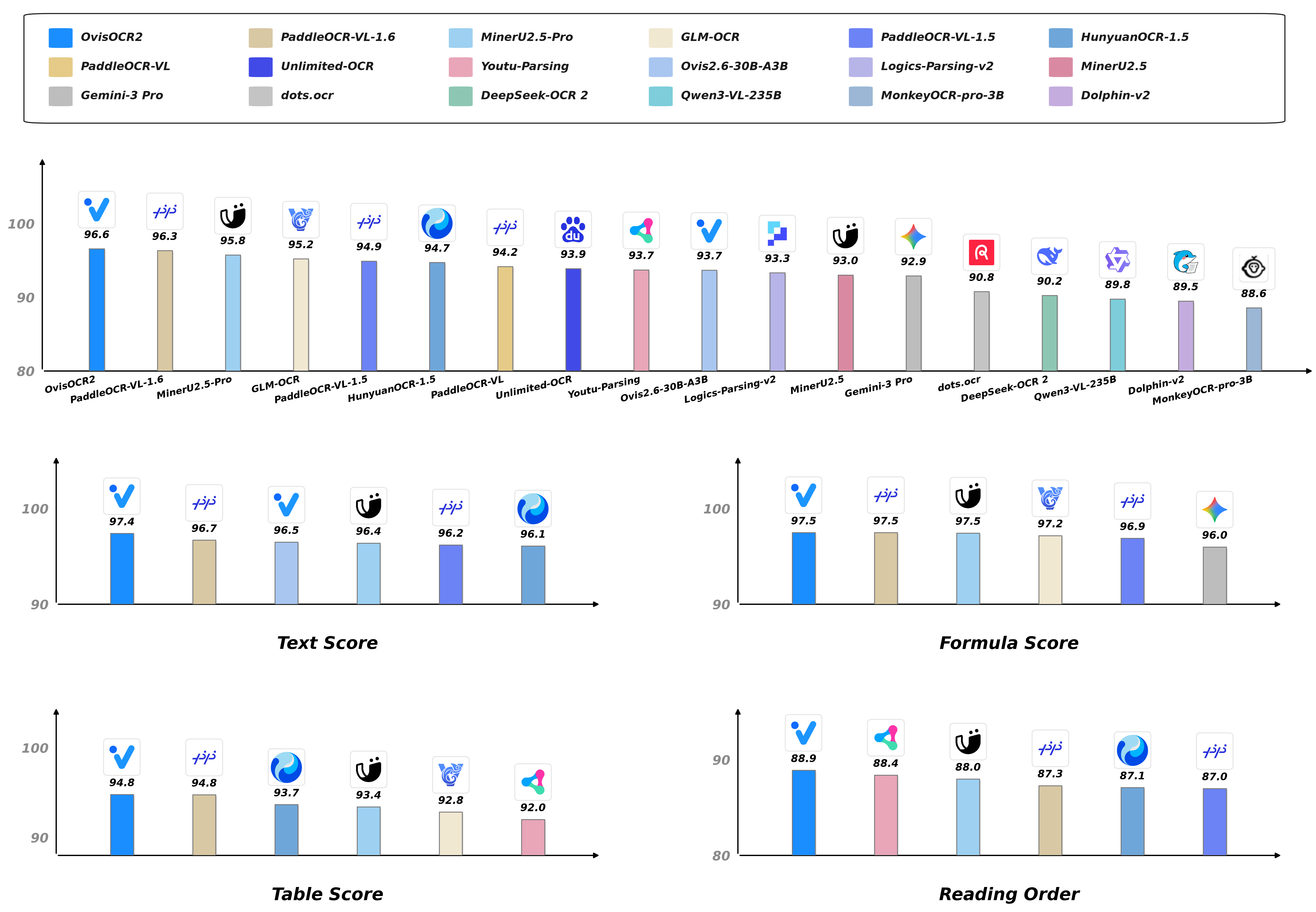}
\caption{Performance of \ours{} on OmniDocBench v1.6}
\label{fig:homepage_performance}
\end{figure}

\section{Introduction}

Document parsing converts visually rich document images into structured, machine-readable representations. Moving beyond plain OCR, the task requires preserving not only textual content but also page organization, reading order, tables, formulas, figures, headers, footers, and other layout-dependent elements. Parsed pages are commonly represented in Markdown, so they can be indexed, retrieved, and used by downstream applications~\cite{omnidocbench,puredocbench,paddleocrvl16,mineru25,glmocr}. We study the page-level image-to-Markdown setting, where a model processes a document image and generates a unified Markdown representation of the page.

Current approaches to this task fall into two families. Pipeline methods decompose a page into layout analysis, region-level content recognition, and page-level merging~\cite{paddleocrvl16,mineru25pro,glmocr}. They remain strong on mainstream document parsing leaderboards such as OmniDocBench v1.6~\cite{omnidocbench,omnidocbenchLeaderboard}, where the top three methods at the time of writing are pipeline-based. This design, however, complicates deployment. Layout parsing and content recognition are usually deployed as separate models, often with different runtime loads. Errors also accumulate across stages: missed table boundaries, imprecise formula crops, or wrong reading-order assignments cannot be fully rectified by a downstream recognizer. End-to-end methods take the opposite route, using a single model to read the document page image and generate the Markdown representation in one pass~\cite{hunyuanocr15,unlimitedocr,jiang2026ovisocr}. The one-pass design simplifies deployment and lets the model condition on page-level context throughout generation. Nevertheless, existing end-to-end methods still lag behind pipeline methods in parsing performance on mainstream benchmarks.

We believe the end-to-end approach is more elegant, and develop \ours{} as a compact document parser by post-training Qwen3.5-0.8B, the smallest model in the Qwen3.5 family~\cite{qwen35}. Our goal is to achieve state-of-the-art document parsing while keeping the deployment footprint small, using a single end-to-end model that surpasses the pipeline methods currently leading public leaderboards. Achieving this with a compact backbone requires both a carefully designed data engine and a training recipe suited to long outputs. The data engine combines real documents with synthetic pages. Real-document annotations are produced by specialized parsers, normalized into a unified Markdown schema, and cleaned through rule-based filtering and subset-level spot-checking. Synthetic page generation starts from HTML templates built from challenging samples; the same HTML source is used to render document images and produce Markdown annotations. Training proceeds through supervised fine-tuning, RL training on a 4B branch, on-policy distillation into the 0.8B model, and model fusion.

We evaluate OvisOCR2 on two public benchmarks and an in-house benchmark. On OmniDocBench v1.6~\cite{omnidocbench,omnidocbenchLeaderboard}, OvisOCR2 reaches an overall score of 96.58, setting a new state of the art and moving an end-to-end parser ahead of the leading pipeline methods on this leaderboard. On PureDocBench~\cite{puredocbench}, OvisOCR2 also ranks first with an Avg3 score of 75.06. To complement these public benchmarks, we construct an in-house benchmark of more than 1,000 pages spanning a broader range of document scenarios. OvisOCR2 achieves the highest overall score on this benchmark and leads across all three difficulty tiers. In particular, for handwriting and complex-table documents, which are both challenging to parse yet common in real-world workflows, OvisOCR2 delivers the strongest overall performance on the corresponding subsets.

\section{Data engine}

\begin{figure}[!t]
\centering
\includegraphics[width=0.88\linewidth]{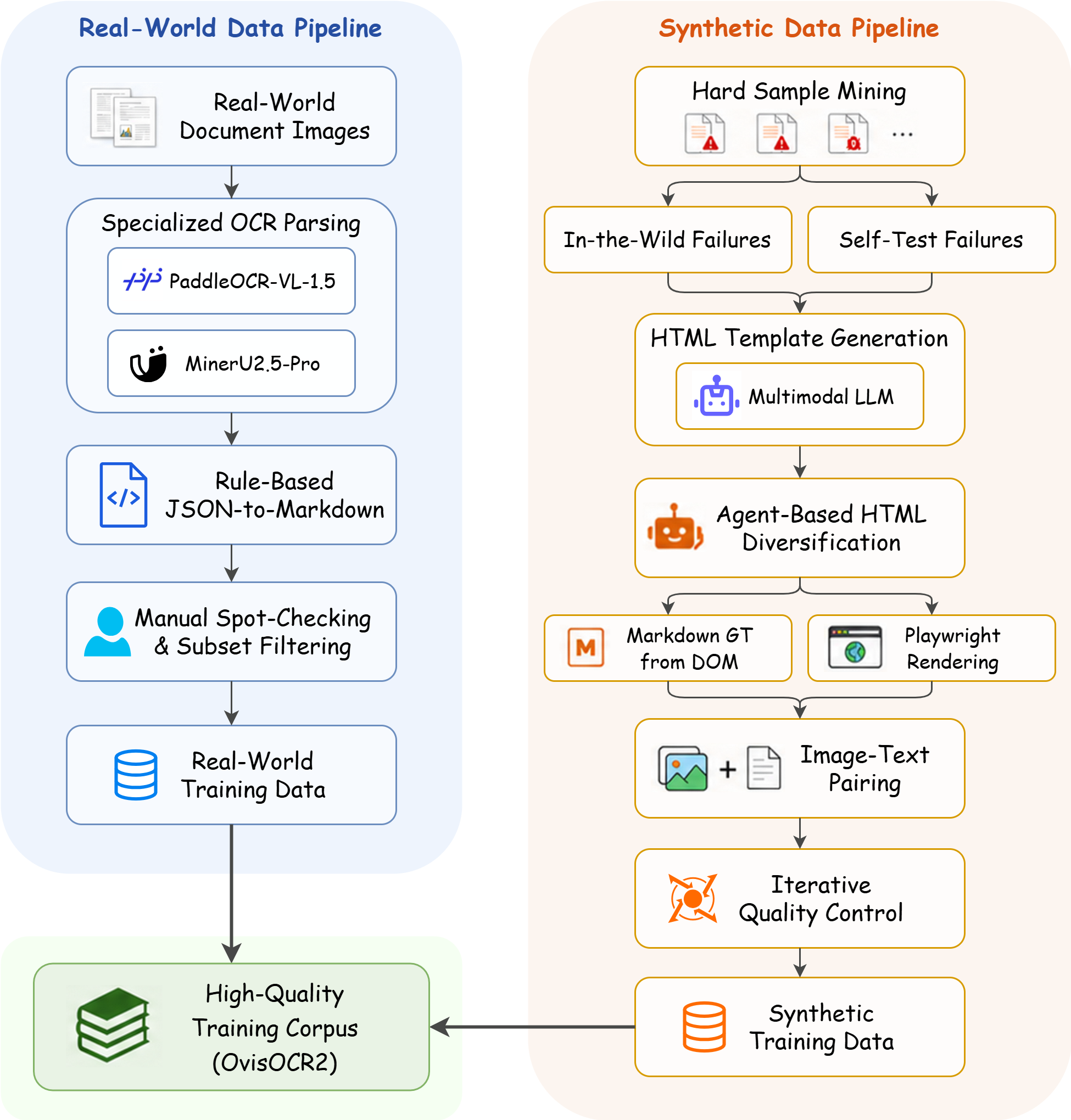}
\caption{Architecture of the data engine}
\label{fig:data_pipeline}
\end{figure}

Training an end-to-end document parser requires supervision that is accurate at multiple levels: fine-grained text transcription, structurally valid serialization, faithful recovery of tables and formulas, and globally coherent reading order. These requirements make data preparation itself a key part of the system design. No single data source is sufficient: real-world documents provide natural layouts and varied visual conditions, but their annotations are noisy and difficult to standardize; synthetic documents offer controllable training targets and scalable coverage, but may suffer from limited realism if synthesized naively. To address this, we build a data engine composed of two complementary pipelines, a real-world data pipeline and a synthetic data pipeline, as shown in Fig.~\ref{fig:data_pipeline}.

\begin{itemize}
    \item \textbf{Real-world data pipeline.}
    This pipeline transforms large-scale real-world document images into reliable document-to-Markdown training data. It exposes the model to real-world visual distributions, including diverse templates, scan qualities, font styles, languages, layout structures, tables, formulas, figures, and other heterogeneous document elements.

    \item \textbf{Synthetic data pipeline.}
    This pipeline complements real-world data by synthesizing controllable document samples with precise ground-truth annotations. Its role is to expand the long-tail coverage of the training distribution, especially for rare or difficult cases that are underrepresented in naturally collected documents, such as table topologies, dense formula-text interleaving, extreme multi-column layouts, and long Markdown outputs.
\end{itemize}

\subsection{Real-world data pipeline}

The real-world pipeline turns document images into training data through conversion and filtering. It first obtains structured outputs from OCR parsers, normalizes them into a common Markdown schema, and then filters the resulting Markdown outputs through rule-based checks and subset-level spot-checking. In this process, parser outputs are used as structured candidates rather than final labels; deterministic conversion and conservative filtering are applied before the data enter the training corpus to improve serialization consistency and structural validity.

\subsubsection{Rule-based structured parsing and Markdown serialization}

For each document image, the real-world data pipeline uses a specialized OCR parser, either PaddleOCR-VL-1.5~\cite{paddleocrvl15} or MinerU2.5-Pro~\cite{mineru25pro}, to obtain structured parsing results. Instead of directly using the raw Markdown strings produced by these systems, we parse their structured JSON responses and convert them into a unified Markdown format through source-specific rules. Each parser output is treated as an ordered list of blocks, where each block contains its category, recognized content, bounding box, and optional parser-specific metadata. The data engine then applies a JSON-to-Markdown converter to normalize these heterogeneous block structures.

The conversion rules are designed around the following principles:

\begin{itemize}
    \item \textbf{Strict category validation.}
    The converter accepts only predefined document categories. MinerU2.5-Pro categories are grouped into text-like elements, tables, visual elements such as images and charts, and container-like blocks that do not directly contribute Markdown content. PaddleOCR-VL-1.5 uses a broader layout vocabulary, covering titles, paragraphs, formulas, tables, figures, headers, footers, seals, footnotes, vertical text, and other document regions. Unknown or unsupported categories are rejected rather than silently ignored, reducing the risk of corrupted annotations entering the training set.

    \item \textbf{Text-block normalization.}
    Text-like blocks with empty content or unreadable placeholders are skipped and recorded by a filtering flag. For MinerU2.5-Pro, adjacent text blocks marked with \texttt{merge\_prev} are merged conservatively: a block is merged only when a valid previous text block exists and the current content is non-empty. Chinese text is concatenated directly, while non-Chinese text is separated with a space. For title blocks, the converter further derives Markdown heading levels from numbering patterns such as decimal section indices, Roman numerals, alphabetic markers, and Chinese section markers, while avoiding over-promotion of year-like numeric titles.

    \item \textbf{Formula normalization.}
    Mathematical expressions are normalized into a unified LaTeX-style Markdown representation. For MinerU2.5-Pro, inline wrappers such as \texttt{\textbackslash(...\textbackslash)} are converted into \texttt{\$...\$}, and display wrappers such as \texttt{\textbackslash[...\textbackslash]} are converted into \texttt{\$\$...\$\$}. Redundant spaces inside math delimiters are removed, line breaks within math wrappers are collapsed, and escaped dollar signs are protected during conversion. Code blocks and inline code spans are excluded from formula rewriting. For PaddleOCR-VL-1.5, the converter applies lightweight post-processing to trim spaces inside single-dollar inline formulas after block concatenation.

    \item \textbf{Table normalization.}
    Table blocks are required to contain valid HTML table structures. Empty tables, non-table contents, malformed table outputs, or repeated tail fragments are filtered out. MinerU2.5-Pro tables are normalized by adding \texttt{border=1} to the \texttt{<table>} tag when the border attribute is absent. PaddleOCR-VL-1.5 tables are sanitized by removing unnecessary HTML style attributes.

    \item \textbf{Visual-region normalization.}
    Visual regions such as figures, charts, header images, and footer images are represented as HTML image tags rather than textual descriptions. Their bounding boxes are normalized to the range $[0,1000)$ and serialized as \\ \texttt{<img src="images/bbox\_left\_top\_right\_bottom.jpg" />}.

    \item \textbf{Parser-specific high-risk cases.}
    Some categories require additional validation. For example, PaddleOCR-VL-1.5 seal and figure-title regions may be wrapped in HTML-like containers; the converter extracts clean text from these containers and rejects empty results. Seal blocks are further checked against layout confidence to remove unreliable detections.
\end{itemize}

After block-level normalization, all valid block contents are concatenated in parser-provided block order using double newlines. The converter also applies pre-filtering checks to remove samples with no valid text-bearing content, duplicated source images, obvious trailing repetitions, or other severe parser-side failures. The resulting Markdown is still not treated as final training data. Rule-based normalization and pre-filtering can remove obvious format and parser failures, but they cannot fully verify whether the content matches the source image. Therefore, we further introduce a spot-checking stage to assess content correspondence and global consistency before retaining the sample for training.

\subsubsection{Manual spot-checking and subset filtering}

We organize candidate datasets according to the data source, parser type, document domain, and conversion configuration. For each dataset subset, we randomly sample document--Markdown pairs and manually compare the converted Markdown with the original document images. The inspection checks whether the serialized Markdown matches the source image along the following dimensions:
\begin{itemize}
    \item \textbf{Text correspondence.}
    We check whether the Markdown output preserves the textual content in the image, without missing text, inserted text, or recognition errors.

    \item \textbf{Formula accuracy.}
    We render formulas according to LaTeX syntax and check whether the rendered results match the formulas in the image.

    \item \textbf{Table alignment.}
    We inspect whether table rows, columns, headers, and cell contents correspond to the original table, especially for dense tables and merged cells.

    \item \textbf{Visual-region alignment.}
    We render the bounding boxes encoded in HTML image tags on the original document image and check whether they cover the intended figures or charts.

    \item \textbf{Reading-order consistency.}
    We examine whether the sampled outputs follow natural human reading order, especially in multi-column pages and pages with dense mixtures of text, formulas, tables, and visual regions.
\end{itemize}

This stage is intentionally conservative. The goal is not to rewrite or repair the converted annotations through free-form generation, but to prevent low-quality data subsets from degrading the training corpus. When a subset contains only occasional minor errors, we retain it together with the sample-level filters applied during rule-based conversion. In contrast, when a subset exhibits frequent errors, it is removed from the training corpus.

\subsection{Synthetic data pipeline}

The synthetic data pipeline follows a source-of-truth principle, generating both the document image and its Markdown target from the same HTML source instead of deriving the target by parsing the rendered image. This design makes the training targets deterministic and avoids parser-derived noise in synthetic labels. The pipeline starts by mining hard samples identified through multi-faceted assessments, uses a multimodal model to convert mined hard samples into initial HTML templates, employs an agent-based generation procedure to expand them into diverse HTML pages, derives Markdown ground truth from the structured source, renders document images with Playwright~\cite{playwright}, and finally packages the resulting image-text pairs into training data after quality inspection.

\subsubsection{Hard sample mining for HTML template generation}

The synthetic data pipeline begins with hard sample mining. Instead of randomly synthesizing generic document pages, we first collect difficult samples from failure cases identified through multi-faceted assessments. These samples are selected when they reveal underrepresented document patterns, such as table-heavy layouts, irregular document structures, header/footer interference, handwritten regions, page-number ambiguity, and complex reading-order patterns.

Each mined hard sample is treated not merely as an isolated failure, but as evidence of a reusable synthesis pattern. We analyze the visual and structural factors behind the failure and group samples with similar typography, table structure, page organization, and localization requirements into the same synthesis family. This grouping allows one template to cover a class of related failure modes rather than overfitting to a single page.

Given a representative hard sample or a cluster of similar hard samples, we use a multimodal model to infer the visual and structural intent of the document. The model converts the observed hard sample into an initial HTML template that preserves the key layout structure needed for subsequent diversification. For elements requiring accurate localization, the template uses explicit DOM spans or element-level wrappers, enabling tight bounding boxes to be recovered during rendering.

The generated HTML template is therefore not a final synthetic sample, but a programmatic abstraction of the mined hard sample distribution. It preserves the challenging factors observed in real failures while exposing controllable variables for subsequent diversification. This hard-sample-centric design converts observed errors into scalable document generators with clean training targets.

\subsubsection{Agent-based HTML diversification}

After obtaining the initial template, we use an agent-based generation procedure to expand the template into a diverse set of HTML pages. The agent edits and extends the HTML template while preserving the visual identity and structural intent of the seed template. This procedure supports iterative code refinement, validity checking, and controlled diversification.

The diversification process introduces randomness at both content and structure levels. Content-level variation covers semantic fields, textual length, numerical values, formulas, and domain-specific terminology. Structure-level variation covers table structure, section hierarchy, page organization, and visual-region placement. Together, these variations expand each seed template into a synthesis family with diverse content and layout configurations.

To keep labels valid during diversification, the agent is constrained by validity rules. Each generated page must remain renderable, visually plausible, category-compliant, and convertible into valid Markdown. The pipeline further maintains domain-specific randomized content pools, allowing synthetic content to vary semantically while keeping the generated labels clean and consistent. This ensures that the synthetic data are large in scale, structurally diverse, and reliably annotated.

\subsubsection{Markdown ground truth generation}

For each diversified HTML page, the data engine generates the Markdown ground truth directly from the corresponding HTML source. The serializer maps source elements into the target Markdown format while preserving their intended structure and reading order, with separate rules for each element type:
\begin{itemize}
    \item \textbf{Text elements.}
    Text nodes, paragraphs, section titles, list items, headers, footers, and footnotes are serialized as standard Markdown text.

    \item \textbf{Tables.}
    Table elements are represented as HTML snippets \texttt{<table>...</table>}, so that both cell content and table structure are retained.

    \item \textbf{Formulas.}
    Mathematical expressions are serialized in LaTeX-style Markdown. Inline formulas and display formulas are represented with consistent math delimiters so that the target format remains compatible with the training format.

    \item \textbf{Visual regions.}
    Each visual region is serialized as an HTML image tag. The coordinates are obtained from the rendered DOM, scaled to $[0,1000)$, and written as \\ \texttt{<img src="images/bbox\_left\_top\_right\_bottom.jpg" />}.
\end{itemize}

Reading order is assigned at the page level. The serializer uses document-type-aware rules rather than a single global spatial heuristic. For regular single-column pages, elements are ordered primarily from top to bottom and secondarily from left to right. For multi-column layouts, the page is partitioned into column regions according to the template layout or the normalized bounding boxes; elements within each column are serialized from top to bottom, and columns are traversed from left to right. For structured elements such as tables, lists, formulas, and anchored visual regions, the serializer preserves their internal DOM order and template-defined associations.

\subsubsection{Document image rendering}

In parallel with Markdown generation, the same HTML page is rendered into a document image. We use Playwright~\cite{playwright} for browser-based rendering, which preserves realistic typography, spacing, table strokes, color styles, wrapping behavior, and page-level layout details. The viewport is automatically adjusted to fit the document canvas, and page boundary handling is applied when necessary to avoid invalid crops, incomplete bottom regions, or mask artifacts.

During rendering, the data engine records element-level geometry from the DOM for elements that require bounding boxes. Explicit spans and element wrappers allow the pipeline to obtain tight bounding boxes for text or visual elements. These boxes are normalized to the target coordinate range and remain consistent with the final screenshot. When visual augmentation is applied, such as mild rotation or other geometric perturbations, the corresponding coordinate transformations are applied to the labels as well.

\subsubsection{Iterative quality control}

Before large-scale generation, we perform a preview-and-iteration procedure. A small batch of samples is first rendered for visual inspection. The inspection focuses on whether the synthesized pages match the intended hard sample style in typography, table layout, page organization, visual-region localization, and reading-order details. If layout defects are observed, the HTML template, agent diversification rules, or rendering parameters are revised before scaling.

After the preview passes inspection, the pipeline expands the randomized content pools and generates samples at scale. The large-scale generation stage supports multi-threaded rendering and writes paired images and labels together with augmented variants when applicable.

The quality control stage removes samples with rendering or pairing failures, empty or degenerate Markdown targets, structural errors, localization errors, or duplicated outputs. Final dataset size is verified across image files, label files, augmented variants, and metadata records before the samples are admitted into the training dataset.

Through this pipeline, we obtain synthetic Markdown training data that are both controllable and precise. The synthetic data pipeline complements the real-world data pipeline by expanding long-tail coverage while preserving clean ground truth and faithful image-text alignment.

\section{Training}

\subsection{Training pipeline}

End-to-end document parsing requires both broad format learning and reliable structured generation. The model must convert a document page image into a Markdown representation while preserving text fidelity, formula renderability, table structure, reading order, and page-level completeness. These requirements call for supervision signals beyond token-level imitation, because many errors in tables, formulas, and long pages are easier to verify structurally than to express through next-token loss alone.

We therefore organize the training pipeline into stages with different roles. Supervised fine-tuning builds the initial end-to-end document parsing policy; reinforcement learning further improves the policy on hard pages using text, formula, and table rewards; on-policy distillation transfers the reward-aligned behavior of a 4B teacher into a deployable 0.8B student; model fusion then forms the final model.

\begin{figure}[!t]
\vspace{-0.3cm}
\centering
\includegraphics[width=\linewidth]{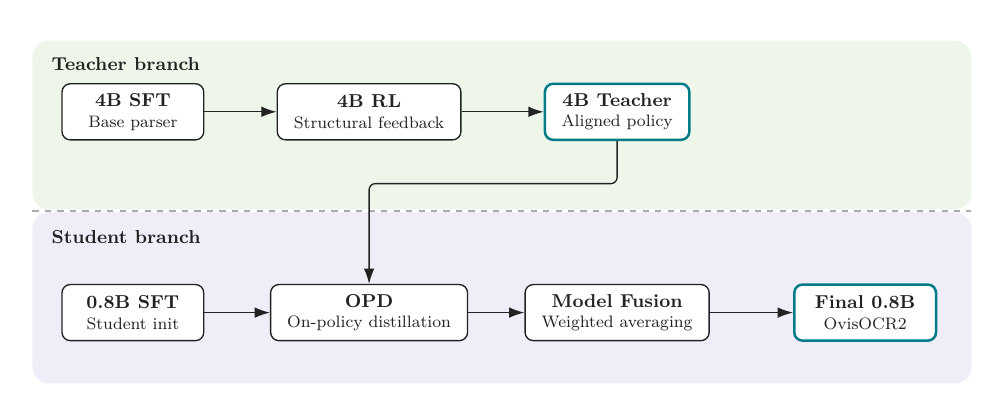}
\caption{Two-branch training of OvisOCR2. The 4B branch produces an RL-aligned teacher, while the 0.8B branch proceeds through SFT, OPD, and model fusion to obtain the final model.}
\end{figure}

\subsection{Supervised fine-tuning}

The SFT stage establishes the base policy for generating Markdown representations from document images. SFT uses the full data mixture described in the data section. The mixture is built on real document pages processed by the real-world data pipeline and is complemented by controlled samples from the synthetic data pipeline. Real data contribute natural layouts, scan quality variations, document types, and visual noise. Synthetic data increase the coverage of tables, formula-intensive pages, long Markdown outputs, and other long-tail structures. This stage therefore provides broad coverage and format consistency for the downstream RL and OPD stages.

We train both the Qwen3.5-0.8B and Qwen3.5-4B models with full-parameter SFT on the same document parsing objective. The 0.8B model is trained for two epochs, whereas the 4B model is trained for 20\% of an epoch to reduce the training cost of the larger branch. The 0.8B SFT checkpoint initializes the deployable student branch, while the 4B SFT checkpoint serves as the base policy for 4B RL; the resulting 4B RL checkpoint is then used as the OPD teacher. We use a 16K maximum sequence length together with a dynamic image-resolution budget, preserving fine-grained visual information without forcing all pages into a fixed resolution.

\subsection{Reinforcement learning}

RL further improves the end-to-end document parsing policy using reward signals on sampled outputs. Many important parsing errors are structural rather than purely lexical: a table may contain mostly correct text while having an incorrect cell topology; a formula may look similar as a string but fail to render or represent a different expression; long pages may suffer from omissions, repetition, or truncation. These errors are weakly expressed by token-level imitation loss, but they can be measured through programmatic checks, rendering-based comparison, and structure-aware parsing.

We use Group Relative Policy Optimization (GRPO)~\cite{shao2024deepseekmath} as the main RL algorithm. For each prompt, we sample multiple responses from the policy and compute group-relative advantages from verifiable rewards, without training an additional value model. This formulation is well suited to document parsing, where responses are long, structured, and evaluable by text, formula, and table analyzers. GRPO trains the model to prefer better candidates among multiple outputs for the same page, thereby reinforcing generation behaviors that are more complete, stable, and structurally valid.

\subsubsection{On-policy hard case construction}

RL data are constructed differently from the broad SFT mixture. The stage primarily uses samples from the synthetic data pipeline, where ground-truth structures are accurate enough to support text, formula, and table rewards. A smaller set of high-quality real documents is retained to preserve natural page distributions and varied visual conditions.

The training set is further shaped through on-policy filtering: we first run the current policy on candidate pages and score the generated responses with the reward function. Very easy and extremely hard samples provide little useful learning signal, and reward-flat samples with little variation across responses are less useful for group-relative learning. Training therefore emphasizes pages on which the model can occasionally produce responses that are clearly better than its average behavior.

\subsubsection{Multi-component reward design}

The RL reward is designed for complete page outputs and avoids reducing document parsing quality to a single text-similarity score. The scalar reward is computed from the components that are actually present in the ground truth: text, tables, and display formulas. Global validity checks, such as max-length truncation and unparseable structures, are applied as guards that zero out the affected available components rather than forming an independent reward term.

Table~\ref{tab:rl_feedback} summarizes the component scores used before page-level aggregation. CDM denotes character detection matching, an image-level metric for formula parsing~\cite{cdm}, and TEDS denotes a tree-edit-distance-based table similarity metric~\cite{pubtabnet}. In the reward, CDM and TEDS are used as normalized scores in \([0,1]\).

\begin{table}[!htbp]
\centering
\small
\setlength{\tabcolsep}{5pt}
\renewcommand{\arraystretch}{1.12}
\caption{Reward components used in RL.}
\label{tab:rl_feedback}
\begin{tabularx}{0.96\linewidth}{@{}>{\raggedright\arraybackslash\bfseries}p{0.22\linewidth}>{\raggedright\arraybackslash}p{0.28\linewidth}Y@{}}
\toprule
\textbf{Component} & \textbf{Score} & \textbf{Measured aspect} \\
\midrule
Text & 1 - normalized edit distance & Text fidelity \\
\addlinespace[2pt]
Formula & CDM & Visual formula matching \\
\addlinespace[2pt]
Table & TEDS & Table content and topology \\
\bottomrule
\end{tabularx}
\end{table}

Let \(\mathcal{C}=\{\mathrm{text},\mathrm{table},\mathrm{formula}\}\). For a page with prediction \(y\) and reference \(y^\ast\), \(s_c(y,y^\ast)\in[0,1]\) denotes the page-level score for component type \(c\). The text score is computed after page-level matching, while formula and table scores are averaged over evaluable ground-truth units after element matching and normalization. Each component has an availability indicator \(a_c(y^\ast)\), which equals one only when the reference contains at least one evaluable unit of that type. The page reward is then computed by averaging only the components available in the reference:

\[
R(y,y^\ast)
=
\frac{
\sum_{c\in\mathcal{C}} a_c(y^\ast)\,s_c(y,y^\ast)
}{
\sum_{c\in\mathcal{C}} a_c(y^\ast)
}.
\]

\subsubsection{Scalable multimodal RL training}

Multimodal RL training for document parsing needs to handle long text responses, page-level image inputs, and rendering-based reward computation. To make the training infra scalable, we apply the following optimizations.

First, reward computation uses hierarchical parallelism and shortcut paths. After rollout, page samples are scored by multiple reward workers in parallel. Within each worker, the formula reward performs matching and normalization first: normalized exact matches are assigned full credit directly, missing or invalid predictions are assigned zero, and only the remaining cases enter CDM rendering comparison. CDM computation is executed through a reusable process pool with per-task timeouts, preventing a small number of problematic formulas from blocking the worker. The table reward follows the same principle: normalized exact matches bypass TEDS, while invalid or unparseable tables receive zero. For long pages or abnormal outputs, the page-level matcher also uses timeout fallback to prevent matching from becoming a tail-latency bottleneck. Reward profiling is kept in the training loop so that slowdowns can be attributed to matching, normalization, rendering-based formula rewards, or table-structure scoring.

Second, the multimodal input transfer path is optimized for high-resolution document pages. Visual preprocessing can produce large page-level tensors, and repeatedly transmitting them across workers would increase host-side memory pressure and communication volume. Training therefore uses object-store-backed reference passing: large visual tensors are stored once and referenced by lightweight handles together with compact metadata. The policy update and reference-constraint computation retrieve and assemble the actual visual inputs when needed. This reduces transfer overhead and prevents high-resolution pages or long batches from moving the training bottleneck to serialization and data movement.

Third, the actor update is optimized for long responses. We use a common-prefix mask so that long shared prefixes among multiple rollouts of the same prompt do not dominate the loss, focusing gradients on tokens after the generation branches diverge. Under the strictly on-policy, single-epoch update setting, the system also avoids an additional old-policy probability pass. These optimizations reduce actor-side computation and memory pressure on long page outputs.

\subsection{On-policy distillation}

\subsubsection{Teacher-guided policy transfer}

A central challenge is to transfer the reward-aligned parsing behavior learned by the 4B RL model into a compact 0.8B model suitable for deployment. The 4B branch has greater capacity and can absorb reward signals more stably, while the compact branch is more sensitive to high-variance updates on long structured outputs. In our preliminary direct-RL comparison, applying the same RL stage directly to the compact model showed larger KL drift and less stable table quality, especially on dense pages and complex layouts.

\begin{figure}[!htbp]
\centering
\includegraphics[width=\linewidth]{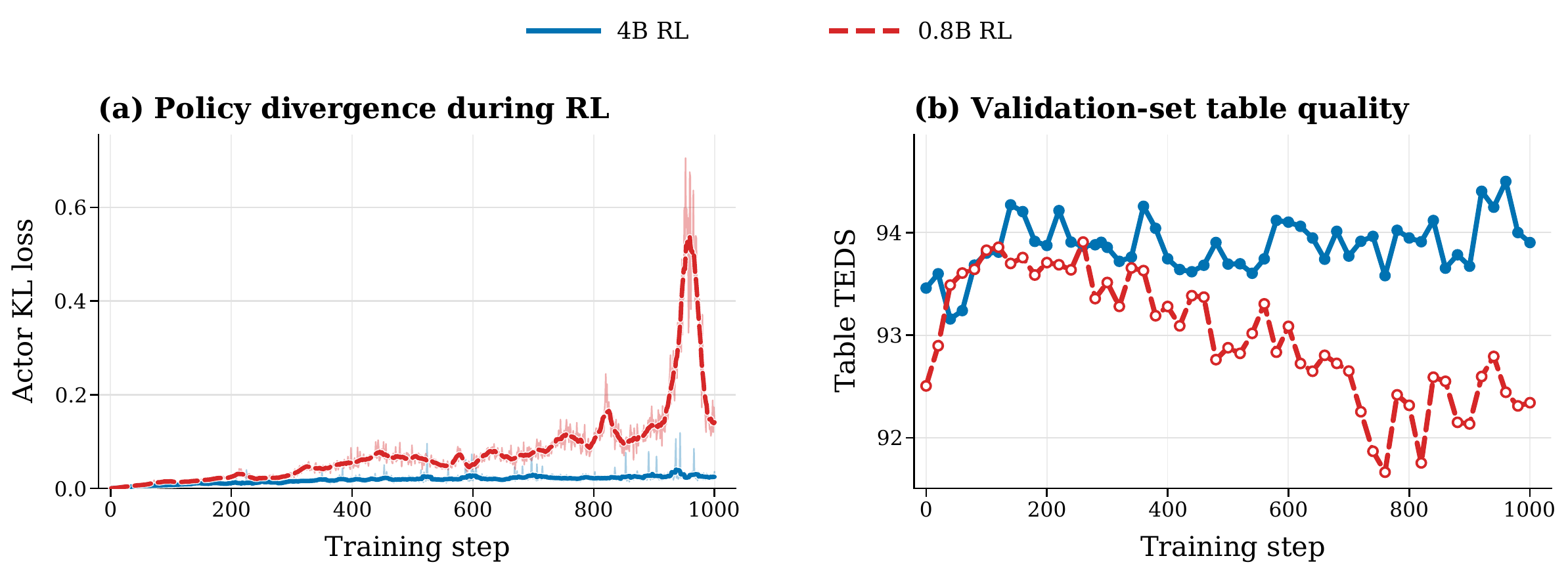}
\caption{Training stability comparison between 4B RL and 0.8B RL. Panel (a) reports actor KL loss, with faint traces showing raw per-step values and bold curves showing 15-step centered rolling means. Panel (b) reports table TEDS on the validation set.}
\label{fig:opd_motivation}
\end{figure}

As shown in Figure~\ref{fig:opd_motivation}, direct 0.8B RL shows higher policy divergence in later training and a corresponding decrease in table quality compared with the 4B RL model. We therefore use the RL-enhanced 4B model as the teacher for the 0.8B student.

The student generates complete page outputs under its current policy, and the teacher evaluates these on-policy trajectories with token-level distribution supervision~\cite{hinton2015distilling,agarwal2024onpolicy}. The compact model thus receives the teacher's parsing preferences for tables, long formulas, and dense layouts through distribution matching, reducing the need to expose the student to another high-variance RL stage.

Formally, given a document image and instruction \(x\), the student policy \(\pi_\theta\) first samples a complete response \(y=(y_1,\ldots,y_T)\). The teacher policy \(\pi_\phi\) then evaluates conditional probabilities for candidate tokens under the same concatenated prompt-response context. Since the teacher and student share the same prefix \(c_t=(x,y_{<t})\), the loss does not require length alignment between teacher and student outputs; supervision is applied directly on states visited by the current student policy.

The main objective is student top-k reverse KL~\cite{li2026rethinkingopd}. At each response position, the support is determined by the current student distribution: \(S_t = \operatorname{TopK}_k\!\left(\pi_\theta(\cdot \mid c_t)\right)\). The teacher only needs to return log probabilities for tokens in \(S_t\). The student and teacher distributions are normalized within this support:

\[
\bar{p}_{t,v}
= \frac{\pi_\theta(v \mid c_t)}{\sum_{u \in S_t}\pi_\theta(u \mid c_t)},
\qquad
\bar{q}_{t,v}
= \frac{\pi_\phi(v \mid c_t)}{\sum_{u \in S_t}\pi_\phi(u \mid c_t)},
\qquad v\in S_t.
\]

Denote the resulting distributions over \(S_t\) by \(\bar{p}_t=(\bar{p}_{t,v})_{v\in S_t}\) and \(\bar{q}_t=(\bar{q}_{t,v})_{v\in S_t}\). Over the set of valid response token positions \(I\), the OPD loss is:

\[
L_{\mathrm{OPD}}(\theta)
=
\frac{1}{|I|}
\sum_{t \in I}
D_{\mathrm{KL}}\!\left(
\bar{p}_t \,\|\, \bar{q}_t
\right).
\]

Here \(I\) denotes token positions used for distillation, excluding prompt and padding tokens. The reverse-KL direction \(D_{\mathrm{KL}}(\bar{p}_t \,\|\, \bar{q}_t)\) gives the objective a mode-seeking behavior: within the compared support, it discourages student probability mass on tokens assigned low probability by the teacher, rather than pushing the student to spread probability over the teacher distribution.

\subsubsection{Efficient OPD training}

Full-vocabulary distillation is expensive for document parsing because page-level responses are long. For a response of length \(T\) and vocabulary size \(V\), matching teacher and student distributions at every response position would require \(T \times V\) log-probability tensors. By restricting the alignment to the student top-\(k\) support \(S_t\), the dominant tensor size is reduced from \(O(TV)\) to \(O(Tk)\).

Teacher scoring is performed on the student-generated trajectory. After student rollout, we keep the top-\(k\) token IDs at each response position and query the teacher for their log probabilities under the same prompt-response prefix. The student update uses the same token IDs. During the student forward pass, only the selected logits required by the OPD loss are gathered, and chunked projection is used for long responses to reduce peak memory.

\subsection{Model fusion}

We train several candidate \ours{} variants by varying both the data mixture and the training configurations. We then apply weighted parameter averaging~\cite{wortsman2022model} for model fusion, producing the final \ours{} model.

\section{Evaluation}

In this section, we assess the effectiveness of \ours{} by comparing against a wide range of document parsing models grouped into general VLMs and specialized VLMs. We further divide specialized VLMs by inference paradigm. The pipeline group includes leading methods such as PaddleOCR-VL-1.6~\cite{paddleocrvl16}, GLM-OCR~\cite{glmocr}, and MinerU2.5-Pro~\cite{mineru25pro}, which combine layout analysis with region-level parsing. The end-to-end group, including \ours{}, outputs the Markdown representation from the page-level image in a single pass. We conduct the evaluation on two public document parsing benchmarks, OmniDocBench v1.6~\cite{omnidocbench} and PureDocBench~\cite{puredocbench}, as well as an in-house benchmark focused on complex document scenarios. We also provide qualitative comparisons in Appendix~\ref{sec:qualitative_examples}.

\subsection{OmniDocBench v1.6}

OmniDocBench v1.6~\cite{omnidocbench} is the most widely adopted public benchmark for page-level document parsing, covering 1,651 PDF pages across 10 document types, 5 layout types, and 5 language types. Compared with the earlier v1.5 protocol, v1.6 refines element matching in formula evaluation and introduces a hard subset of structurally complex pages, yielding more discriminative rankings among strong methods. The benchmark jointly evaluates four dimensions: text transcription (measured by normalized edit distance), formula recognition (character detection matching, CDM~\cite{cdm}), table reconstruction (tree-edit similarity, TEDS and TEDS-S~\cite{pubtabnet}), and reading-order recovery (edit distance over block sequences). The overall score is computed by averaging the text score converted from normalized edit distance, formula CDM, and table TEDS.

As shown in Table~\ref{tab:odb16_full}, on OmniDocBench v1.6, \ours{} achieves a state-of-the-art overall score of 96.58 with a compact 0.8B model. It surpasses leading pipeline methods, including PaddleOCR-VL-1.6, MinerU2.5-Pro, and GLM-OCR, and improves over the previous best end-to-end method by 1.84 points. Across metrics, \ours{} reports the lowest text edit distance, the highest formula CDM score, the tied highest TEDS score, the highest TEDS-S score, and the lowest reading-order edit distance.

\begin{table}[!t]
\centering
\caption{OmniDocBench v1.6 comparison. Public leaderboard scores are from OpenDataLab~\cite{omnidocbenchLeaderboard}; PaddleOCR-VL-1.6, Unlimited-OCR, and HunyuanOCR-1.5 are from their technical reports~\cite{paddleocrvl16,unlimitedocr,hunyuanocr15}. RO denotes reading order.}
\label{tab:odb16_full}
\resizebox{\linewidth}{!}{
\begin{tabular}{@{}l|lc|cccccc@{}}
\toprule
\textbf{Model type} & \textbf{Method} & \textbf{Parameters} & \textbf{Overall}\up & \textbf{Text$^{Edit}$}\down & \textbf{Formula$^{CDM}$}\up & \textbf{Table$^{TEDS}$}\up & \textbf{Table$^{TEDS\mbox{-}S}$}\up & \textbf{RO$^{Edit}$}\down \\
\midrule
\multirow{7}{*}{\makecell[l]{\textbf{General}\\\textbf{VLMs}}}
& InternVL3.5-241B~\cite{internvl} & 241B & 83.76 & 0.130 & 89.95 & 74.35 & 79.78 & 0.215 \\
& Kimi K2.5~\cite{kimiteam2026kimik25visualagentic} & 1T/32B & 84.53 & 0.107 & 83.50 & 80.76 & 84.00 & 0.211 \\
& GPT-5.2~\cite{gpt52} & -- & 86.59 & 0.114 & 88.21 & 82.95 & 87.93 & 0.193 \\
& Qwen3-VL-235B~\cite{qwen3vl235b} & 235B/22B & 89.78 & 0.063 & 92.55 & 83.07 & 86.75 & 0.166 \\
& Gemini 3 Flash~\cite{gemini3flash} & -- & 92.62 & 0.066 & 95.16 & 89.29 & 93.51 & 0.172 \\
& Gemini 3 Pro~\cite{gemini3pro} & -- & 92.91 & 0.064 & 95.99 & 89.15 & 92.96 & 0.165 \\
& Ovis2.6-30B-A3B~\cite{ovis26} & 30B/3B & 93.70 & 0.035 & 95.17 & 89.44 & 92.40 & 0.135 \\
\midrule
\multirow{10}{*}{\makecell[l]{\textbf{Specialized VLMs}\\\textbf{(pipeline)}}}
& Dolphin-1.5~\cite{dolphin} & 0.3B & 86.52 & 0.094 & 87.49 & 81.43 & 84.82 & 0.167 \\
& MonkeyOCR-pro-3B~\cite{monkeyocr} & 3B & 88.57 & 0.074 & 88.74 & 84.35 & 88.62 & 0.189 \\
& Dolphin-v2~\cite{dolphin} & 3B & 89.50 & 0.069 & 91.01 & 84.40 & 87.44 & 0.150 \\
& MinerU2.5~\cite{mineru25} & 1.2B & 93.04 & 0.045 & 95.77 & 87.88 & 91.47 & 0.130 \\
& Youtu-Parsing~\cite{youtu-parsing} & 2.5B & 93.74 & 0.044 & 93.63 & 92.02 & 95.00 & \secondcell{0.116} \\
& PaddleOCR-VL~\cite{paddleocrvl} & 0.9B & 94.18 & 0.040 & 95.91 & 90.65 & 93.74 & 0.135 \\
& PaddleOCR-VL-1.5~\cite{paddleocrvl15} & 0.9B & 94.93 & 0.038 & 96.89 & 91.67 & 94.37 & 0.130 \\
& GLM-OCR~\cite{glmocr} & 0.9B & 95.22 & 0.044 & 97.18 & 92.83 & 95.39 & 0.133 \\
& MinerU2.5-Pro~\cite{mineru25pro} & 1.2B & 95.75 & 0.036 & 97.45 & 93.42 & 95.92 & 0.120 \\
& PaddleOCR-VL-1.6~\cite{paddleocrvl16} & 0.9B & \secondcell{96.33} & \secondcell{0.033} & \secondcell{97.49} & \bestcell{94.76} & \secondcell{97.11} & 0.127 \\
\midrule
\multirow{15}{*}{\makecell[l]{\textbf{Specialized VLMs}\\\textbf{(end-to-end)}}}
& POINTS-Reader~\cite{pointsreader} & 3B & 83.37 & 0.096 & 85.72 & 73.98 & 77.40 & 0.198 \\
& Nanonets-OCR-S~\cite{Nanonets-OCR-S} & 3B & 83.61 & 0.108 & 81.46 & 80.18 & 84.51 & 0.213 \\
& olmOCR~\cite{olmocr} & 7B & 85.74 & 0.139 & 88.10 & 83.00 & 87.17 & 0.216 \\
& OCRVerse~\cite{ocrverse} & 4B & 88.60 & 0.063 & 89.61 & 82.44 & 86.27 & 0.163 \\
& HunyuanOCR~\cite{hunyuanocr} & 1B & 89.95 & 0.088 & 87.68 & 91.01 & 93.23 & 0.171 \\
& DeepSeek-OCR-2~\cite{deepseekocr2} & 3B/0.5B & 90.25 & 0.050 & 91.84 & 83.89 & 87.75 & 0.144 \\
& OpenDoc-0.1B~\cite{unirec} & 0.1B & 90.67 & 0.049 & 93.02 & 83.88 & 87.45 & 0.140 \\
& dots.ocr~\cite{dots-ocr} & 3B & 90.77 & 0.048 & 89.95 & 87.18 & 90.58 & 0.138 \\
& FireRed-OCR~\cite{fireredocr} & 2B & 93.26 & 0.037 & 95.44 & 88.04 & 91.06 & 0.131 \\
& ABot-OCR~\cite{abotocr} & 2B & 93.30 & 0.037 & 94.86 & 88.69 & 91.87 & 0.137 \\
& Logics-Parsing-v2~\cite{logicsparsing} & 4B & 93.33 & 0.041 & 95.65 & 88.42 & 91.98 & 0.137 \\
& Qianfan-OCR~\cite{qianfanocr} & 4B & 93.90 & 0.040 & 95.08 & 90.53 & 93.31 & 0.130 \\
& Unlimited-OCR~\cite{unlimitedocr} & 3B/0.5B & 93.92 & 0.042 & 95.79 & 90.16 & 93.32 & 0.129 \\
& HunyuanOCR-1.5~\cite{hunyuanocr15} & 1B & 94.74 & 0.039 & 94.50 & \secondcell{93.67} & 94.71 & 0.129 \\
& \textbf{\ours} & \textbf{0.8B} & \bestcell{96.58} & \bestcell{0.025} & \bestcell{97.53} & \bestcell{94.76} & \bestcell{97.16} & \bestcell{0.111} \\
\bottomrule
\end{tabular}
}
\end{table}

\subsection{PureDocBench}

PureDocBench~\cite{puredocbench} complements OmniDocBench with source-traceable evaluation data: document images are rendered from HTML sources, and annotations are produced from the same source rather than transcribed from rendered images. It evaluates robustness across the Clean track with rendered pages, the Digital track with degraded images, and the Real track with physical or screen-mediated recaptures, including phone-captured pages, photocopies, screen photography, and compressed screenshots, covering 1,475 pages across 10 domains and 66 subcategories, for 4,425 images in total.

\begin{table}[!htbp]
\centering
\caption{PureDocBench comparison. The Clean, Digital, and Real columns report per-track overall scores; Avg3 is their mean. Public scores are transcribed from the PureDocBench main table~\cite{puredocbench}.}
\label{tab:puredoc_full}
\renewcommand{\arraystretch}{0.86}
\resizebox{\linewidth}{!}{
\begin{tabular}{@{}l|lc|cccc@{}}
\toprule
\textbf{Model type} & \textbf{Method} & \textbf{Parameters} & \textbf{Clean}\up & \textbf{Digital}\up & \textbf{Real}\up & \textbf{Avg3}\up \\
\midrule
\multirow{6}{*}{\makecell[l]{\textbf{General}\\\textbf{VLMs}}}
& MiniCPM-V-4.5~\cite{minicpmv} & 8B & 51.81 & 49.38 & 37.59 & 46.26 \\
& Step3-VL~\cite{step3vl} & 10B & 53.65 & 52.74 & 45.06 & 50.48 \\
& Kimi K2.6~\cite{kimiteam2026kimik25visualagentic} & 1T/32B & 72.32 & 69.95 & 68.02 & 70.10 \\
& Gemini-3.1-Pro~\cite{gemini31pro} & -- & 70.04 & 69.28 & \bestcell{71.98} & 70.43 \\
& Qwen3.5-9B~\cite{qwen35} & 9B & 73.87 & 73.34 & 65.45 & 70.89 \\
& Qwen3.5-122B-A10B~\cite{qwen35} & 122B/10B & 76.14 & \secondcell{76.34} & \secondcell{69.85} & \secondcell{74.11} \\
\midrule
\multirow{10}{*}{\makecell[l]{\textbf{Specialized VLMs}\\\textbf{(pipeline)}}}
& OpenOCR~\cite{svtrv2,unirec} & 0.1B & 32.70 & 30.03 & 25.73 & 29.49 \\
& MonkeyOCR-pro-1.2B~\cite{monkeyocr} & 1.2B & 61.09 & 55.72 & 43.82 & 53.54 \\
& MonkeyOCR-pro-3B~\cite{monkeyocr} & 3B & 62.23 & 57.40 & 46.49 & 55.37 \\
& Dolphin-v2~\cite{dolphin} & 3B & 65.90 & 60.24 & 44.92 & 57.02 \\
& GLM-OCR~\cite{glmocr} & 0.9B & 68.65 & 63.06 & 58.31 & 63.34 \\
& PaddleOCR-VL-1.5~\cite{paddleocrvl15} & 0.9B & 73.01 & 66.73 & 60.50 & 66.75 \\
& MinerU2.5~\cite{mineru25} & 1.2B & 74.90 & 68.92 & 59.15 & 67.66 \\
& Youtu-Parsing~\cite{youtu-parsing} & 2B & 75.02 & 69.66 & 60.29 & 68.32 \\
& MinerU2.5-Pro~\cite{mineru25pro} & 1.2B & 75.87 & 71.77 & 62.56 & 70.07 \\
& DotsMOCR~\cite{dots-mocr} & 3B & 76.27 & 73.16 & 61.73 & 70.39 \\
\midrule
\multirow{16}{*}{\makecell[l]{\textbf{Specialized VLMs}\\\textbf{(end-to-end)}}}
& OCRFlux-3B~\cite{ocrflux} & 3B & 47.14 & 41.82 & 37.21 & 42.06 \\
& DeepSeek-OCR~\cite{deepseekocr} & 3B/0.5B & 53.50 & 46.95 & 40.48 & 46.98 \\
& UniRec-0.1B~\cite{unirec} & 0.1B & 58.91 & 52.42 & 34.44 & 48.59 \\
& DeepSeek-OCR-2~\cite{deepseekocr2} & 3B/0.5B & 55.53 & 49.41 & 43.60 & 49.51 \\
& Qianfan-OCR~\cite{qianfanocr} & 4B & 57.22 & 50.85 & 45.06 & 51.04 \\
& OpenDoc-0.1B~\cite{unirec} & 0.1B & 60.28 & 52.46 & 44.27 & 52.00 \\
& olmOCR~\cite{olmocr} & 7B & 62.56 & 57.84 & 47.30 & 55.90 \\
& Nanonets-OCR2~\cite{Nanonets-OCR2} & 3B & 64.83 & 61.23 & 49.03 & 58.36 \\
& HunyuanOCR~\cite{hunyuanocr} & 1B & 65.61 & 61.49 & 54.58 & 60.56 \\
& olmOCR-2-7B~\cite{olmocr2} & 7B & 69.36 & 65.87 & 56.10 & 63.78 \\
& dots.ocr~\cite{dots-ocr} & 3B & 72.01 & 65.95 & 55.68 & 64.55 \\
& FireRed-OCR~\cite{fireredocr} & 2B & 70.81 & 68.49 & 57.42 & 65.57 \\
& OCRVerse~\cite{ocrverse} & 4B & 73.18 & 71.36 & 63.66 & 69.40 \\
& Logics-Parsing-v2~\cite{logicsparsing} & 4B & 76.35 & 73.85 & 67.64 & 72.61 \\
& FD-RL~\cite{fdrl} & 4B & \secondcell{78.38} & 76.33 & 67.04 & 73.92 \\
& \textbf{\ours} & \textbf{0.8B} & \bestcell{81.55} & \bestcell{77.09} & 66.56 & \bestcell{75.06} \\
\bottomrule
\end{tabular}
}
\end{table}

As shown in Table~\ref{tab:puredoc_full}, \ours{} achieves a state-of-the-art Avg3 score of 75.06. It also ranks first on the clean and digital tracks. On the real track, \ours{} remains below strong general VLMs such as Gemini-3.1-Pro and Qwen3.5-122B-A10B, suggesting that robustness to degraded real-world images remains an important direction for future work.

\subsection{In-house benchmark}

\begin{table}[!t]
\centering
\caption{Results on the in-house benchmark.}
\label{tab:inhouse_all}
\small
\setlength{\tabcolsep}{2pt}
\begin{tabular*}{\linewidth}{@{\extracolsep{\fill}}lcccccc@{}}
\toprule
Model & Text$^{Edit}$\down & Formula$^{CDM}$\up & Table$^{TEDS}$\up & Table$^{TEDS\mbox{-}S}$\up & RO$^{Edit}$\down & Overall\up \\
\midrule
MinerU2.5-Pro~\cite{mineru25pro} & 0.1645 & 34.85 & \secondcell{78.20} & \secondcell{82.68} & 0.3351 & 65.54 \\
PaddleOCR-VL-1.5~\cite{paddleocrvl15} & 0.1325 & 84.73 & 73.18 & 77.84 & 0.2380 & 81.55 \\
GLM-OCR~\cite{glmocr} & 0.1378 & 84.04 & 78.12 & 82.25 & 0.2393 & 82.80 \\
PaddleOCR-VL-1.6~\cite{paddleocrvl16} & \secondcell{0.1292} & \secondcell{85.13} & 76.42 & 80.74 & \secondcell{0.2358} & \secondcell{82.88} \\
\textbf{\ours} & \bestcell{0.0850} & \bestcell{86.32} & \bestcell{78.80} & \bestcell{82.87} & \bestcell{0.1885} & \bestcell{85.54} \\
\bottomrule
\end{tabular*}
\end{table}

\begin{table}[!t]
\centering
\caption{Overall performance on the in-house benchmark by difficulty level.}
\label{tab:inhouse_all_difficulty}
\small
\setlength{\tabcolsep}{2pt}
\begin{tabular*}{\linewidth}{@{\extracolsep{\fill}}lccc@{}}
\toprule
Model & Easy\up & Medium\up & Hard\up \\
\midrule
MinerU2.5-Pro~\cite{mineru25pro} & 66.12 & 65.37 & 64.60 \\
PaddleOCR-VL-1.5~\cite{paddleocrvl15} & 83.23 & 82.23 & 71.07 \\
GLM-OCR~\cite{glmocr} & 84.94 & \secondcell{83.31} & 72.61 \\
PaddleOCR-VL-1.6~\cite{paddleocrvl16} & \secondcell{84.98} & 83.16 & \secondcell{75.06} \\
\textbf{\ours} & \bestcell{87.95} & \bestcell{85.82} & \bestcell{78.99} \\
\bottomrule
\end{tabular*}
\end{table}

\begin{table}[!t]
\centering
\caption{Results on the handwriting subset of the in-house benchmark.}
\label{tab:inhouse_handwriting}
\small
\setlength{\tabcolsep}{2pt}
\begin{tabular*}{\linewidth}{@{\extracolsep{\fill}}lcccccc@{}}
\toprule
Model & Overall\up & Text$^{Edit}$\down & Formula$^{CDM}$\up & Table$^{TEDS}$\up & Table$^{TEDS\mbox{-}S}$\up & RO$^{Edit}$\down \\
\midrule
MinerU2.5-Pro~\cite{mineru25pro} & 55.04 & 0.2903 & 40.63 & \secondcell{53.52} & \bestcell{67.21} & 0.4353 \\
PaddleOCR-VL-1.5~\cite{paddleocrvl15} & 67.31 & 0.2257 & 77.24 & 47.27 & 57.74 & 0.2283 \\
PaddleOCR-VL-1.6~\cite{paddleocrvl16} & 67.66 & \secondcell{0.2148} & \secondcell{77.88} & 46.57 & 57.39 & \secondcell{0.2202} \\
GLM-OCR~\cite{glmocr} & \secondcell{69.58} & 0.2421 & 75.63 & \bestcell{57.31} & \secondcell{65.76} & 0.2340 \\
\textbf{\ours} & \bestcell{72.28} & \bestcell{0.1561} & \bestcell{81.51} & 50.95 & 59.84 & \bestcell{0.1733} \\
\bottomrule
\end{tabular*}
\end{table}

\begin{table}[!t]
\centering
\caption{Results on the complex-table subset of the in-house benchmark. MR denotes missing rate, defined as the fraction of tables missing from the parsed output.}
\label{tab:inhouse_table}
\small
\setlength{\tabcolsep}{2pt}
\begin{tabular*}{\linewidth}{@{\extracolsep{\fill}}lccccc@{}}
\toprule
Model & Overall\up & Text$^{Edit}$\down & Table$^{TEDS}$\up & Table$^{TEDS\mbox{-}S}$\up & MR\down \\
\midrule
PaddleOCR-VL-1.5~\cite{paddleocrvl15} & 70.98 & \secondcell{0.2705} & 69.01 & 73.24 & 0.1637 \\
PaddleOCR-VL-1.6~\cite{paddleocrvl16} & 73.19 & 0.2760 & 73.98 & 77.41 & 0.1593 \\
GLM-OCR~\cite{glmocr} & \secondcell{74.08} & 0.2848 & 76.63 & \secondcell{79.92} & 0.1726 \\
MinerU2.5-Pro~\cite{mineru25pro} & 71.43 & 0.3407 & \secondcell{76.92} & 79.71 & \secondcell{0.1327} \\
\textbf{\ours} & \bestcell{83.97} & \bestcell{0.1040} & \bestcell{78.34} & \bestcell{81.33} & \bestcell{0.0796} \\
\bottomrule
\end{tabular*}
\end{table}

Although existing public benchmarks cover a broad range of document layouts, their data distributions do not fully reflect the heterogeneity and visual noise of documents encountered in real-world workflows. Examples include, among others, domain-specific forms, scanned reports with seals, printed templates with handwritten annotations, and tables with irregularly merged cells. To provide a more comprehensive evaluation of \ours{}, we construct an in-house benchmark following the OmniDocBench v1.6 evaluation protocol. The benchmark comprises more than 1,000 pages spanning a wide range of document types.

During the development of \ours{}, we found handwriting and complex tables to be two challenging document scenarios frequently encountered in real-world workflows. We therefore report separate results on handwriting and complex-table subsets in addition to the full benchmark.

As shown in Table~\ref{tab:inhouse_all}, \ours{} achieves the highest overall score on the in-house benchmark and shows consistently leading performance across text, formula, table, and reading-order metrics. We further divide the benchmark by parsing difficulty. As shown in Table~\ref{tab:inhouse_all_difficulty}, the advantage of \ours{} holds across easy, medium, and hard tiers.

Tables~\ref{tab:inhouse_handwriting} and~\ref{tab:inhouse_table} summarize the handwriting and complex-table results. On handwriting, while GLM-OCR obtains a higher table TEDS score, \ours{} achieves the highest overall score, the lowest text edit distance, the highest formula CDM score, and the lowest reading-order error. On the complex-table subset, \ours{} obtains the highest overall score, the highest TEDS, and the lowest missing rate. The missing-rate comparison is particularly informative: pipeline methods miss 13--17\% of tables during layout parsing, an error that is irrecoverable by downstream recognizers.

\section{Conclusion}

In this report, we present \ours{}, a compact 0.8B end-to-end model for page-level document parsing. We build a data engine that combines filtered real-document annotations with source-aligned synthetic pages, and use a training recipe that applies supervised fine-tuning, reinforcement learning, on-policy distillation, and model fusion. Evaluated on both public and in-house benchmarks, \ours{} establishes new state-of-the-art results on OmniDocBench v1.6 and PureDocBench, while also achieving the highest overall score in our in-house comparison. Future work will focus on improving robustness to degraded real-world images and further strengthening performance on handwriting-heavy and complex-table documents.

\section*{Author List}
Shiyin Lu, Yinglun Li, Yu Xia, Yuhui Chen, An-Yang Ji, Jun-Peng Jiang, Qing-Guo Chen,
Jianshan Zhao, En Lin, Haijun Li, Cheng Qin, Zhao Xu, Weihua Luo.

\bibliographystyle{unsrt}
\bibliography{references}

\begin{thebibliography}{10}

\bibitem{omnidocbench}
Linke Ouyang, Yuan Qu, Hongbin Zhou, Jiawei Zhu, Rui Zhang, Qunshu Lin, Bin
  Wang, Zhiyuan Zhao, Man Jiang, Xiaomeng Zhao, Jin Shi, Fan Wu, Pei Chu,
  Minghao Liu, Zhenxiang Li, Chao Xu, Bo~Zhang, Botian Shi, Zhongying Tu, and
  Conghui He.
\newblock {OmniDocBench}: Benchmarking diverse {PDF} document parsing with
  comprehensive annotations.
\newblock In {\em Proceedings of the IEEE/CVF Conference on Computer Vision and
  Pattern Recognition (CVPR)}, pages 24838--24848, June 2025.

\bibitem{puredocbench}
Zhiheng Li, Zongyang Ma, Jiaxian Chen, Jianing Zhang, Zhaolong Su, Yutong
  Zhang, Zhiyin Yu, Ruiqi Liu, Xiaolei Lv, Bo~Li, Jun Gao, Ziqi Zhang, Chunfeng
  Yuan, Bing Li, and Weiming Hu.
\newblock How far is document parsing from solved? {PureDocBench}: A
  source-traceable benchmark across clean, degraded, and real-world settings.
\newblock {\em arXiv preprint arXiv:2605.07492}, 2026.

\bibitem{paddleocrvl16}
Zelun Zhang, Hongen Liu, Suyin Liang, Yubo Zhang, Yiqing Xiang, Jiaxuan Liu,
  Ting Sun, Manhui Lin, Yue Zhang, Changda Zhou, Tingquan Gao, Cheng Cui,
  Yi~Liu, Dianhai Yu, and Yanjun Ma.
\newblock {PaddleOCR-VL-1.6}: Expanding the frontier of document parsing with
  under-optimized region refinement and progressive post-training.
\newblock {\em arXiv preprint arXiv:2606.03264}, 2026.

\bibitem{mineru25}
Junbo Niu, Zheng Liu, Zhuangcheng Gu, Bin Wang, Linke Ouyang, Zhiyuan Zhao, Tao
  Chu, Tianyao He, Fan Wu, Qintong Zhang, Zhenjiang Jin, Guang Liang, Rui
  Zhang, Wenzheng Zhang, Yuan Qu, Zhifei Ren, Yuefeng Sun, Zirui Tang, Boyu
  Niu, Yuanhong Zheng, Dongsheng Ma, Ziyang Miao, Hejun Dong, Siyi Qian,
  Junyuan Zhang, Fangdong Wang, Jingzhou Chen, Xiaomeng Zhao, Liqun Wei, Wei
  Li, Shasha Wang, RuiLiang Xu, Yuanyuan Cao, Lu~Chen, Qianqian Wu, Huaiyu Gu,
  Lindong Lu, Dechen Lin, Shenguanlin, Xuanhe Zhou, Linfeng Zhang, Yuhang Zang,
  Xiaoyi Dong, Jiaqi Wang, Bo~Zhang, Lei Bai, Pei Chu, Weijia Li, Jiang Wu,
  Lijun Wu, Zhenxiang Li, Guangyu Wang, Zhongying Tu, Chao Xu, Kai Chen, Bowen
  Zhou, Dahua Lin, Wentao Zhang, and Conghui He.
\newblock {MinerU2.5}: A decoupled vision-language model for efficient
  high-resolution document parsing.
\newblock In {\em Proceedings of the 64th Annual Meeting of the Association for
  Computational Linguistics (Volume 6: Industry Track)}, pages 13--42, San
  Diego, California, USA, July 2026. Association for Computational Linguistics.

\bibitem{glmocr}
Shuaiqi Duan, Yadong Xue, Weihan Wang, Zhe Su, Huan Liu, Sheng Yang, Guobing
  Gan, Guo Wang, Zihan Wang, Shengdong Yan, Dexin Jin, Yuxuan Zhang, Guohong
  Wen, Yanfeng Wang, Yutao Zhang, Xiaohan Zhang, Wenyi Hong, Yukuo Cen, Da~Yin,
  Bin Chen, Wenmeng Yu, Xiaotao Gu, and Jie Tang.
\newblock {GLM-OCR} technical report.
\newblock {\em arXiv preprint arXiv:2603.10910}, 2026.

\bibitem{mineru25pro}
Bin Wang, Tianyao He, Linke Ouyang, Fan Wu, Zhiyuan Zhao, Tao Chu, Yuan Qu,
  Zhenjiang Jin, Weijun Zeng, Ziyang Miao, Bangrui Xu, Junbo Niu, Mengzhang
  Cai, Jiantao Qiu, Qintong Zhang, Dongsheng Ma, Yuefeng Sun, Hejun Dong,
  Wenzheng Zhang, Jutao Xiao, Jiayong Shi, Pengyu Liao, Xiaomeng Zhao, Huaping
  Zhong, Liqun Wei, Jing Yu, Jie Yang, Wei Li, Shasha Wang, Qianqian Wu, Xuanhe
  Zhou, Weijia Li, Zhenxiang Li, Zhongying Tu, Jiang Wu, Lijun Wu, Chao Xu, Kai
  Chen, Wentao Zhang, Yu~Qiao, Bowen Zhou, Dahua Lin, and Conghui He.
\newblock {MinerU2.5-Pro}: Pushing the limits of data-centric document parsing
  at scale.
\newblock {\em arXiv preprint arXiv:2604.04771}, 2026.

\bibitem{omnidocbenchLeaderboard}
{OpenDataLab}.
\newblock {OmniDocBench} public leaderboard.
\newblock \url{https://opendatalab.com/omnidocbench}, 2026.
\newblock Accessed June 29, 2026.

\bibitem{hunyuanocr15}
Gengluo Li, Xingyu Wan, Shangpin Peng, Weinong Wang, Hao Feng, Yongkun Du,
  Binghong Wu, Zheng Ruan, Zhiqiong Lu, Liang Wu, Pengyuan Lyu, Huawen Shen,
  Zibin Lin, Shijing Hu, Jieneng Yang, Hongbing Wen, Guanghua Yu, Hong Liu,
  Bochao Wang, Can Ma, Han Hu, Chengquan Zhang, and Yu~Zhou.
\newblock {HunyuanOCR-1.5}: Making lightweight {OCR} {VLMs} faster and better.
\newblock {\em arXiv preprint arXiv:2607.04884}, 2026.

\bibitem{unlimitedocr}
Youyang Yin, Huanhuan Liu, YY, Qunyi Xie, Chaorun Liu, Shiqi Yang, Shaohua
  Wang, Zhanlong Liu, Hao Zou, Jinyue Chen, Shu Wei, Jingjing Wu, Mingxin
  Huang, Zhen Wu, Guibin Wang, Tengyu Du, and Lei Jia.
\newblock Unlimited {OCR} works.
\newblock {\em arXiv preprint arXiv:2606.23050}, 2026.

\bibitem{jiang2026ovisocr}
Jun-Peng Jiang, Shiyin Lu, An-Yang Ji, Yinglun Li, Qing-Guo Chen, Zhao Xu,
  Weihua Luo, Kaifu Zhang, De-Chuan Zhan, and Han-Jia Ye.
\newblock {OvisOCR}: End-to-end document parsing via aligning specialized
  perception with general reasoning.
\newblock In {\em Proceedings of the 43rd International Conference on Machine
  Learning}, volume 306 of {\em Proceedings of Machine Learning Research},
  Seoul, South Korea, 2026. PMLR.

\bibitem{qwen35}
{Qwen Team}.
\newblock {Qwen3.5}: Towards native multimodal agents.
\newblock \url{https://qwen.ai/blog?id=qwen3.5}, 2026.

\bibitem{paddleocrvl15}
Cheng Cui, Ting Sun, Suyin Liang, Tingquan Gao, Zelun Zhang, Jiaxuan Liu,
  Xueqing Wang, Changda Zhou, Hongen Liu, Manhui Lin, Yue Zhang, Yubo Zhang,
  Yi~Liu, Dianhai Yu, and Yanjun Ma.
\newblock {PaddleOCR-VL-1.5}: Towards a multi-task {0.9B} {VLM} for robust
  in-the-wild document parsing.
\newblock {\em arXiv preprint arXiv:2601.21957}, 2026.

\bibitem{playwright}
{Microsoft}.
\newblock Playwright: Fast and reliable end-to-end testing for modern web apps.
\newblock \url{https://playwright.dev/}, 2026.

\bibitem{shao2024deepseekmath}
Zhihong Shao, Peiyi Wang, Qihao Zhu, Runxin Xu, Junxiao Song, Xiao Bi, Haowei
  Zhang, Mingchuan Zhang, Y.~K. Li, Y.~Wu, and Daya Guo.
\newblock {DeepSeekMath}: Pushing the limits of mathematical reasoning in open
  language models.
\newblock {\em arXiv preprint arXiv:2402.03300}, 2024.

\bibitem{cdm}
Bin Wang, Fan Wu, Linke Ouyang, Zhuangcheng Gu, Rui Zhang, Renqiu Xia, Botian
  Shi, Bo~Zhang, and Conghui He.
\newblock Image over text: Transforming formula recognition evaluation with
  character detection matching.
\newblock In {\em Proceedings of the IEEE/CVF Conference on Computer Vision and
  Pattern Recognition (CVPR)}, 2025.

\bibitem{pubtabnet}
Xu~Zhong, Elaheh ShafieiBavani, and Antonio~Jimeno Yepes.
\newblock Image-based table recognition: Data, model, and evaluation.
\newblock In {\em Proceedings of the European Conference on Computer Vision
  (ECCV)}, 2020.

\bibitem{hinton2015distilling}
Geoffrey Hinton, Oriol Vinyals, and Jeff Dean.
\newblock Distilling the knowledge in a neural network.
\newblock {\em arXiv preprint arXiv:1503.02531}, 2015.

\bibitem{agarwal2024onpolicy}
Rishabh Agarwal, Nino Vieillard, Yongchao Zhou, Piotr Stanczyk, Sabela Ramos,
  Matthieu Geist, and Olivier Bachem.
\newblock On-policy distillation of language models: Learning from
  self-generated mistakes.
\newblock In {\em International Conference on Learning Representations}, 2024.

\bibitem{li2026rethinkingopd}
Yaxuan Li, Yuxin Zuo, Bingxiang He, Jinqian Zhang, Chaojun Xiao, Cheng Qian,
  Tianyu Yu, Huan-ang Gao, Wenkai Yang, Zhiyuan Liu, and Ning Ding.
\newblock Rethinking on-policy distillation of large language models:
  Phenomenology, mechanism, and recipe.
\newblock {\em arXiv preprint arXiv:2604.13016}, 2026.

\bibitem{wortsman2022model}
Mitchell Wortsman, Gabriel Ilharco, Samir~Yitzhak Gadre, Rebecca Roelofs,
  Raphael Gontijo-Lopes, Ari~S. Morcos, Hongseok Namkoong, Ali Farhadi, Yair
  Carmon, Simon Kornblith, and Ludwig Schmidt.
\newblock {Model Soups}: Averaging weights of multiple fine-tuned models
  improves accuracy without increasing inference time.
\newblock In {\em Proceedings of the 39th International Conference on Machine
  Learning}, pages 23965--23998, 2022.

\bibitem{internvl}
Weiyun Wang, Zhangwei Gao, Lixin Gu, Hengjun Pu, Long Cui, Xingguang Wei,
  Zhaoyang Liu, Linglin Jing, Shenglong Ye, Jie Shao, Zhaokai Wang, Zhe Chen,
  Hongjie Zhang, Ganlin Yang, Haomin Wang, Qi~Wei, Jinhui Yin, Wenhao Li, Erfei
  Cui, Guanzhou Chen, Zichen Ding, Changyao Tian, Zhenyu Wu, Jingjing Xie,
  Zehao Li, Bowen Yang, Yuchen Duan, Xuehui Wang, Zhi Hou, Haoran Hao, Tianyi
  Zhang, Songze Li, Xiangyu Zhao, Haodong Duan, Nianchen Deng, Bin Fu, Yinan
  He, Yi~Wang, Conghui He, Botian Shi, Junjun He, Yingtong Xiong, Han Lv, Lijun
  Wu, Wenqi Shao, Kaipeng Zhang, Huipeng Deng, Biqing Qi, Jiaye Ge, Qipeng Guo,
  Wenwei Zhang, Songyang Zhang, Maosong Cao, Junyao Lin, Kexian Tang, Jianfei
  Gao, Haian Huang, Yuzhe Gu, Chengqi Lyu, Huanze Tang, Rui Wang, Haijun Lv,
  Wanli Ouyang, Limin Wang, Min Dou, Xizhou Zhu, Tong Lu, Dahua Lin, Jifeng
  Dai, Weijie Su, Bowen Zhou, Kai Chen, Yu~Qiao, Wenhai Wang, and Gen Luo.
\newblock {InternVL3.5}: Advancing open-source multimodal models in
  versatility, reasoning, and efficiency.
\newblock {\em arXiv preprint arXiv:2508.18265}, 2025.

\bibitem{kimiteam2026kimik25visualagentic}
{Kimi Team}.
\newblock {Kimi K2.5}: Visual agentic intelligence.
\newblock {\em arXiv preprint arXiv:2602.02276}, 2026.

\bibitem{gpt52}
{OpenAI}.
\newblock Update to {GPT-5} system card: {GPT-5.2}.
\newblock
  \url{https://cdn.openai.com/pdf/3a4153c8-c748-4b71-8e31-aecbde944f8d/oai_5_2_system-card.pdf},
  2025.

\bibitem{qwen3vl235b}
Shuai Bai, Yuxuan Cai, Ruizhe Chen, Keqin Chen, Xionghui Chen, Zesen Cheng,
  Lianghao Deng, Wei Ding, Chang Gao, Chunjiang Ge, Wenbin Ge, Zhifang Guo,
  Qidong Huang, Jie Huang, Fei Huang, Binyuan Hui, Shutong Jiang, Zhaohai Li,
  Mingsheng Li, Mei Li, Kaixin Li, Zicheng Lin, Junyang Lin, Xuejing Liu,
  Jiawei Liu, Chenglong Liu, Yang Liu, Dayiheng Liu, Shixuan Liu, Dunjie Lu,
  Ruilin Luo, Chenxu Lv, Rui Men, Lingchen Meng, Xuancheng Ren, Xingzhang Ren,
  Sibo Song, Yuchong Sun, Jun Tang, Jianhong Tu, Jianqiang Wan, Peng Wang,
  Pengfei Wang, Qiuyue Wang, Yuxuan Wang, Tianbao Xie, Yiheng Xu, Haiyang Xu,
  Jin Xu, Zhibo Yang, Mingkun Yang, Jianxin Yang, An~Yang, Bowen Yu, Fei Zhang,
  Hang Zhang, Xi~Zhang, Bo~Zheng, Humen Zhong, Jingren Zhou, Fan Zhou, Jing
  Zhou, Yuanzhi Zhu, and Ke~Zhu.
\newblock {Qwen3-VL} technical report.
\newblock {\em arXiv preprint arXiv:2511.21631}, 2025.

\bibitem{gemini3flash}
{Google DeepMind}.
\newblock {Gemini 3 Flash} model card.
\newblock \url{https://deepmind.google/models/model-cards/gemini-3-flash/},
  2025.

\bibitem{gemini3pro}
{Google DeepMind}.
\newblock {Gemini 3 Pro} model card.
\newblock \url{https://deepmind.google/models/model-cards/gemini-3-pro/}, 2025.

\bibitem{ovis26}
Shiyin Lu, Yang Li, Yu~Xia, Yuwei Hu, Shanshan Zhao, Yanqing Ma, Zhichao Wei,
  Yinglun Li, Lunhao Duan, Jianshan Zhao, Yuxuan Han, Haijun Li, Wanying Chen,
  Junke Tang, Chengkun Hou, Zhixing Du, Tianli Zhou, Wenjie Zhang, Huping Ding,
  Jiahe Li, Wen Li, Gui Hu, Yiliang Gu, Siran Yang, Jiamang Wang, Hailong Sun,
  Yibo Wang, Hui Sun, Jinlong Huang, Yuping He, Shengze Shi, Weihong Zhang,
  Guodong Zheng, Junpeng Jiang, Sensen Gao, Yi-Feng Wu, Sijia Chen, Yuhui Chen,
  Qing-Guo Chen, Zhao Xu, Weihua Luo, and Kaifu Zhang.
\newblock {Ovis2.5 Technical Report}.
\newblock {\em arXiv:2508.11737}, 2025.

\bibitem{dolphin}
Hao Feng, Shu Wei, Xiang Fei, Wei Shi, Yingdong Han, Lei Liao, Jinghui Lu,
  Binghong Wu, Qi~Liu, Chunhui Lin, Jingqun Tang, Hao Liu, and Can Huang.
\newblock Dolphin: Document image parsing via heterogeneous anchor prompting.
\newblock In {\em Findings of the Association for Computational Linguistics:
  ACL 2025}, pages 21919--21936, Vienna, Austria, July 2025. Association for
  Computational Linguistics.

\bibitem{monkeyocr}
Zhang Li, Yuliang Liu, Qiang Liu, Zhiyin Ma, Ziyang Zhang, Shuo Zhang, Biao
  Yang, Zidun Guo, Jiarui Zhang, Xinyu Wang, and Xiang Bai.
\newblock {MonkeyOCR}: Document parsing with a structure-recognition-relation
  triplet paradigm.
\newblock {\em arXiv preprint arXiv:2506.05218}, 2025.

\bibitem{youtu-parsing}
Kun Yin, Yunfei Wu, Bing Liu, Zhongpeng Cai, Xiaotian Li, Huang Chen, Xin Li,
  Haoyu Cao, Yinsong Liu, Deqiang Jiang, Xing Sun, Yunsheng Wu, Qianyu Li,
  Antai Guo, Yanzhen Liao, Yanqiu Qu, Haodong Lin, Chengxu He, and Shuangyin
  Liu.
\newblock {Youtu-Parsing}: Perception, structuring and recognition via
  high-parallelism decoding.
\newblock {\em arXiv preprint arXiv:2601.20430}, 2026.

\bibitem{paddleocrvl}
Cheng Cui, Ting Sun, Suyin Liang, Tingquan Gao, Zelun Zhang, Jiaxuan Liu,
  Xueqing Wang, Changda Zhou, Hongen Liu, Manhui Lin, Yue Zhang, Yubo Zhang,
  Handong Zheng, Jing Zhang, Jun Zhang, Yi~Liu, Dianhai Yu, and Yanjun Ma.
\newblock {PaddleOCR-VL}: Boosting multilingual document parsing via a {0.9B}
  ultra-compact vision-language model.
\newblock {\em arXiv preprint arXiv:2510.14528}, 2025.

\bibitem{pointsreader}
Yuan Liu, Zhongyin Zhao, Le~Tian, Haicheng Wang, Xubing Ye, Yangxiu You, Zilin
  Yu, Chuhan Wu, Xiao Zhou, Yang Yu, and Jie Zhou.
\newblock {POINTS-Reader}: Distillation-free adaptation of vision-language
  models for document conversion.
\newblock In {\em Proceedings of the Conference on Empirical Methods in Natural
  Language Processing (EMNLP)}, 2025.

\bibitem{Nanonets-OCR-S}
Souvik Mandal, Ashish Talewar, Paras Ahuja, and Prathamesh Juvatkar.
\newblock {Nanonets-OCR-S}: A model for transforming documents into structured
  {Markdown} with intelligent content recognition and semantic tagging.
\newblock \url{https://nanonets.com/research/nanonets-ocr-s}, 2025.

\bibitem{olmocr}
Jake Poznanski, Aman Rangapur, Jon Borchardt, Jason Dunkelberger, Regan Huff,
  Daniel Lin, Christopher Wilhelm, Kyle Lo, and Luca Soldaini.
\newblock {olmOCR}: Unlocking trillions of tokens in {PDFs} with vision
  language models.
\newblock {\em arXiv preprint arXiv:2502.18443}, 2025.

\bibitem{ocrverse}
Yufeng Zhong, Lei Chen, Xuanle Zhao, Wenkang Han, Liming Zheng, Jing Huang,
  Deyang Jiang, Yilin Cao, Lin Ma, and Zhixiong Zeng.
\newblock {OCRVerse}: Towards holistic {OCR} in end-to-end vision-language
  models.
\newblock {\em arXiv preprint arXiv:2601.21639}, 2026.

\bibitem{hunyuanocr}
{Hunyuan Vision Team}, Pengyuan Lyu, Xingyu Wan, Gengluo Li, Shangpin Peng,
  Weinong Wang, Liang Wu, Huawen Shen, Yu~Zhou, Canhui Tang, Qi~Yang, Qiming
  Peng, Bin Luo, Hower Yang, Xinsong Zhang, Jinnian Zhang, Houwen Peng,
  Hongming Yang, Senhao Xie, Longsha Zhou, Ge~Pei, Binghong Wu, Rui Yan, Kan
  Wu, Jieneng Yang, Bochao Wang, Kai Liu, Jianchen Zhu, Jie Jiang, Linus, Han
  Hu, and Chengquan Zhang.
\newblock {HunyuanOCR} technical report.
\newblock {\em arXiv preprint arXiv:2511.19575}, 2025.

\bibitem{deepseekocr2}
Haoran Wei, Yaofeng Sun, and Yukun Li.
\newblock {DeepSeek-OCR} 2: Visual causal flow.
\newblock {\em arXiv preprint arXiv:2601.20552}, 2026.

\bibitem{unirec}
Yongkun Du, Zhineng Chen, Yazhen Xie, Weikang Bai, Hao Feng, Wei Shi, Yuchen
  Su, Can Huang, and Yu-Gang Jiang.
\newblock {UniRec-0.1B}: Unified text and formula recognition with {0.1B}
  parameters.
\newblock {\em arXiv preprint arXiv:2512.21095}, 2025.

\bibitem{dots-ocr}
Yumeng Li, Guang Yang, Hao Liu, Bowen Wang, and Colin Zhang.
\newblock dots.ocr: Multilingual document layout parsing in a single
  vision-language model.
\newblock {\em arXiv preprint arXiv:2512.02498}, 2025.

\bibitem{fireredocr}
Hao Wu, Haoran Lou, Xinyue Li, Zuodong Zhong, Zhaojun Sun, Phellon Chen, Xuanhe
  Zhou, Kai Zuo, Yibo Chen, Xu~Tang, Yao Hu, Boxiang Zhou, Jian Wu, Yongji Wu,
  Wenxin Yu, Yingmiao Liu, Yuhao Huang, Manjie Xu, Gang Liu, Yidong Ma, Zhichao
  Sun, and Changhao Qiao.
\newblock {FireRed-OCR} technical report.
\newblock {\em arXiv preprint arXiv:2603.01840}, 2026.

\bibitem{abotocr}
Kaitao Jiang, Ruiyan Gong, Xiaolong Cheng, Kangning Niu, Tianlun Li, and Mu~Xu.
\newblock {ABot-OCR} technical report.
\newblock {\em arXiv preprint arXiv:2605.27978}, 2026.

\bibitem{logicsparsing}
Xiangyang Chen, Shuzhao Li, Xiuwen Zhu, Yongfan Chen, Fan Yang, Cheng Fang, Lin
  Qu, Xiaoxiao Xu, Hu~Wei, and Minggang Wu.
\newblock {Logics-Parsing} technical report.
\newblock {\em arXiv preprint arXiv:2509.19760}, 2025.

\bibitem{qianfanocr}
Daxiang Dong, Mingming Zheng, Dong Xu, Chunhua Luo, Bairong Zhuang, Yuxuan Li,
  Ruoyun He, Haoran Wang, Wenyu Zhang, Wenbo Wang, Yicheng Wang, Xue Xiong,
  Ayong Zheng, Xiaoying Zuo, Ziwei Ou, Jingnan Gu, Quanhao Guo, Jianmin Wu,
  Dawei Yin, and Dou Shen.
\newblock {Qianfan-OCR}: A unified end-to-end model for document intelligence.
\newblock {\em arXiv preprint arXiv:2603.13398}, 2026.

\bibitem{minicpmv}
Tianyu Yu, Zefan Wang, Chongyi Wang, Fuwei Huang, Wenshuo Ma, Zhihui He,
  Tianchi Cai, Weize Chen, Yuxiang Huang, Yuanqian Zhao, Bokai Xu, Junbo Cui,
  Yingjing Xu, Liqing Ruan, Luoyuan Zhang, Hanyu Liu, Jingkun Tang, Hongyuan
  Liu, Qining Guo, Wenhao Hu, Bingxiang He, Jie Zhou, Jie Cai, Ji~Qi, Zonghao
  Guo, Chi Chen, Guoyang Zeng, Yuxuan Li, Ganqu Cui, Ning Ding, Xu~Han, Yuan
  Yao, Zhiyuan Liu, and Maosong Sun.
\newblock {MiniCPM-V 4.5}: Cooking efficient {MLLMs} via architecture, data,
  and training recipe.
\newblock {\em arXiv preprint arXiv:2509.18154}, 2025.

\bibitem{step3vl}
Ailin Huang, Chengyuan Yao, Chunrui Han, Fanqi Wan, Hangyu Guo, Haoran Lv,
  Hongyu Zhou, Jia Wang, Jian Zhou, Jianjian Sun, Jingcheng Hu, Kangheng Lin,
  Liang Zhao, Mitt Huang, Song Yuan, Wenwen Qu, Xiangfeng Wang, Yanlin Lai,
  Yingxiu Zhao, Yinmin Zhang, Yukang Shi, Yuyang Chen, Zejia Weng, Ziyang Meng,
  Ang Li, Aobo Kong, Bo~Dong, Changyi Wan, David Wang, Di~Qi, Dingming Li,
  En~Yu, Guopeng Li, Haiquan Yin, Han Zhou, Hanshan Zhang, Haolong Yan, Hebin
  Zhou, Hongbo Peng, Jiaran Zhang, Jiashu Lv, Jiayi Fu, Jie Cheng, Jie Zhou,
  Jisheng Yin, Jingjing Xie, Jingwei Wu, Jun Zhang, Junfeng Liu, Kaijun Tan,
  Kaiwen Yan, Liangyu Chen, Lina Chen, Mingliang Li, Qian Zhao, Quan Sun,
  Shaoliang Pang, Shengjie Fan, Shijie Shang, Siyuan Zhang, Tianhao You, Wei
  Ji, Wuxun Xie, Xiaobo Yang, Xiaojie Hou, Xiaoran Jiao, Xiaoxiao Ren, Xiangwen
  Kong, Xin Huang, Xin Wu, Xing Chen, Xinran Wang, Xuelin Zhang, Yana Wei, Yang
  Li, Yanming Xu, Yeqing Shen, Yuang Peng, Yue Peng, Yu~Zhou, Yusheng Li,
  Yuxiang Yang, Yuyang Zhang, Zhe Xie, Zhewei Huang, Zhenyi Lu, Zhimin Fan,
  Zihui Cheng, Daxin Jiang, Qi~Han, Xiangyu Zhang, Yibo Zhu, and Zheng Ge.
\newblock {STEP3-VL-10B} technical report.
\newblock {\em arXiv preprint arXiv:2601.09668}, 2026.

\bibitem{gemini31pro}
{Google DeepMind}.
\newblock {Gemini 3.1 Pro} model card.
\newblock \url{https://deepmind.google/models/model-cards/gemini-3-1-pro/},
  2026.

\bibitem{svtrv2}
Yongkun Du, Zhineng Chen, Hongtao Xie, Caiyan Jia, and Yu-Gang Jiang.
\newblock {SVTRv2}: {CTC} beats encoder-decoder models in scene text
  recognition.
\newblock In {\em Proceedings of the IEEE/CVF International Conference on
  Computer Vision (ICCV)}, pages 20147--20156, 2025.

\bibitem{dots-mocr}
Handong Zheng, Yumeng Li, Kaile Zhang, Liang Xin, Guangwei Zhao, Hao Liu, Jiayu
  Chen, Jie Lou, Qi~Fu, Rui Yang, Shuo Jiang, Weijian Luo, Weijie Su, Weijun
  Zhang, Xingyu Zhu, Yabin Li, Yiwei Ma, Yu~Chen, Yuqiu Ji, Zhaohui Yu, Guang
  Yang, Colin Zhang, Lei Zhang, Yuliang Liu, and Xiang Bai.
\newblock Multimodal {OCR}: Parse anything from documents.
\newblock {\em arXiv preprint arXiv:2603.13032}, 2026.

\bibitem{ocrflux}
{ChatDOC}.
\newblock {OCRFlux}: A multimodal toolkit for converting documents into
  {Markdown}.
\newblock \url{https://github.com/chatdoc-com/OCRFlux}, 2025.

\bibitem{deepseekocr}
Haoran Wei, Yaofeng Sun, and Yukun Li.
\newblock {DeepSeek-OCR}: Contexts optical compression.
\newblock {\em arXiv preprint arXiv:2510.18234}, 2025.

\bibitem{Nanonets-OCR2}
Souvik Mandal, Ashish Talewar, Siddhant Thakuria, Paras Ahuja, and Prathamesh
  Juvatkar.
\newblock {Nanonets-OCR2}: A model for transforming documents into structured
  {Markdown} with intelligent content recognition and semantic tagging.
\newblock \url{https://nanonets.com/research/nanonets-ocr-2}, 2025.

\bibitem{olmocr2}
Jake Poznanski, Luca Soldaini, and Kyle Lo.
\newblock {olmOCR} 2: Unit test rewards for document {OCR}.
\newblock {\em arXiv preprint arXiv:2510.19817}, 2025.

\bibitem{fdrl}
Yufeng Zhong, Lei Chen, Zhixiong Zeng, Xuanle Zhao, Deyang Jiang, Liming Zheng,
  Jing Huang, Haibo Qiu, Peng Shi, Siqi Yang, and Lin Ma.
\newblock Reading or reasoning? format decoupled reinforcement learning for
  document {OCR}.
\newblock In {\em Proceedings of the IEEE/CVF Conference on Computer Vision and
  Pattern Recognition (CVPR)}, 2026.

\end{thebibliography}

\clearpage
\appendix
\section{Qualitative examples}
\label{sec:qualitative_examples}

This appendix presents qualitative comparisons on table and handwritten documents.
Red highlights mark issues such as text insertions, recognition errors, and table-structure errors,
while omitted text is shown in light gray within square brackets.

\setcounter{figure}{0}
\renewcommand{\thefigure}{\thesection.\arabic{figure}}

\begin{figure}[!htbp]
\centering
\includegraphics[width=0.86\linewidth]{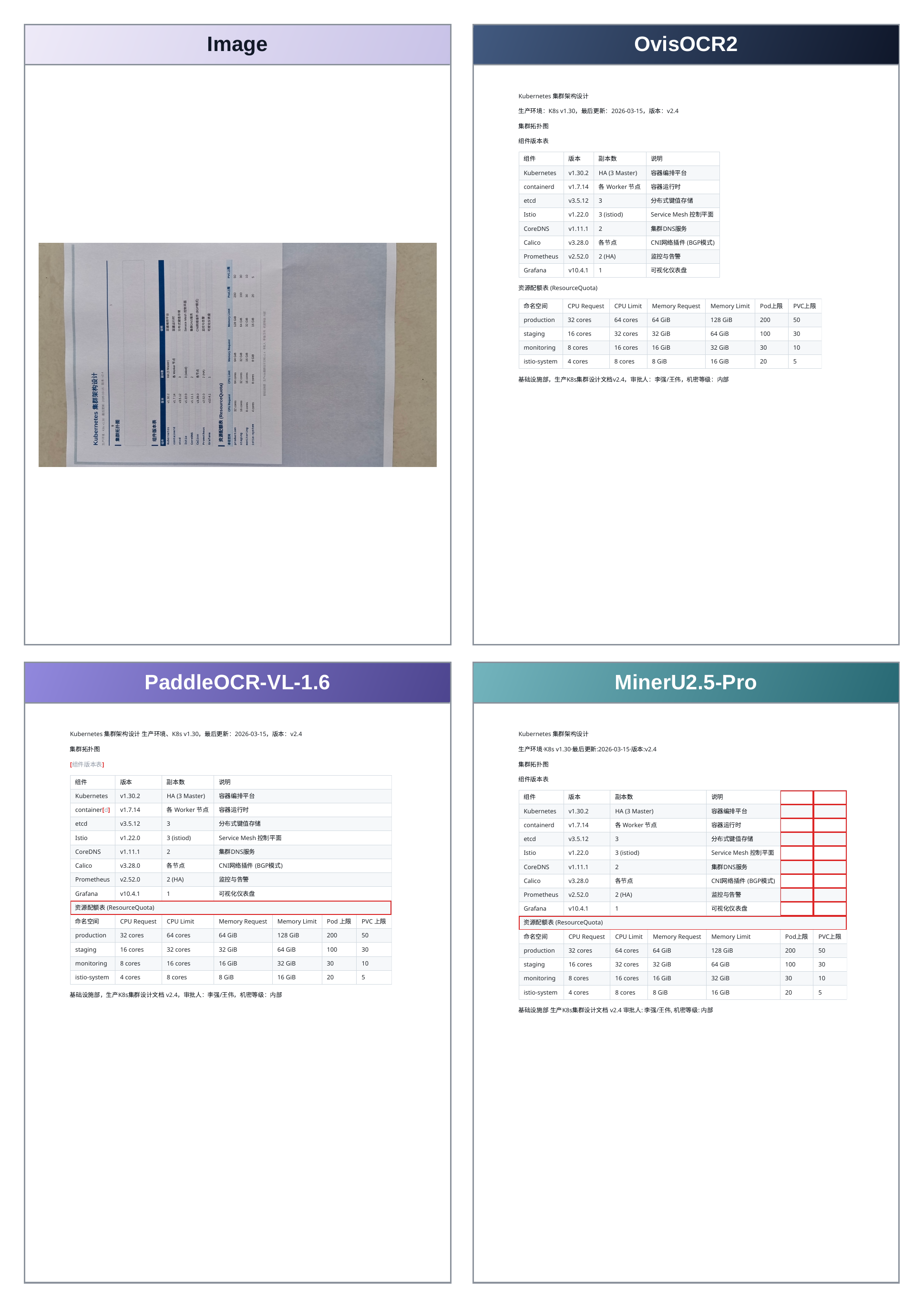}
\caption{Qualitative comparison on a table document.}
\label{fig:caset1}
\end{figure}

\clearpage
\begin{figure}[!htbp]
\ContinuedFloat
\centering
\includegraphics[width=\linewidth]{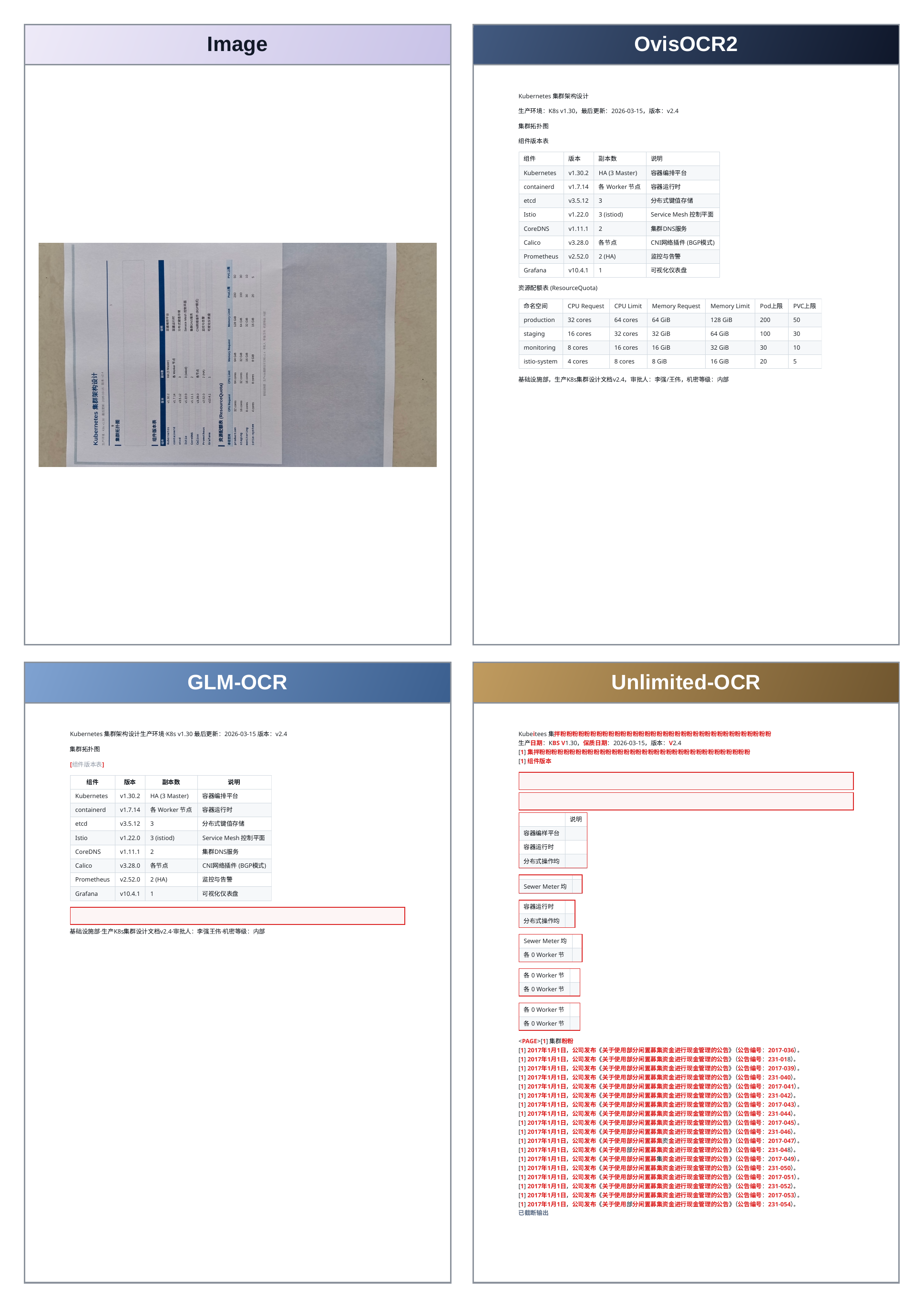}
\caption{Qualitative comparison on a table document (continued).}
\end{figure}

\clearpage
\begin{figure}[!htbp]
\centering
\includegraphics[width=\linewidth]{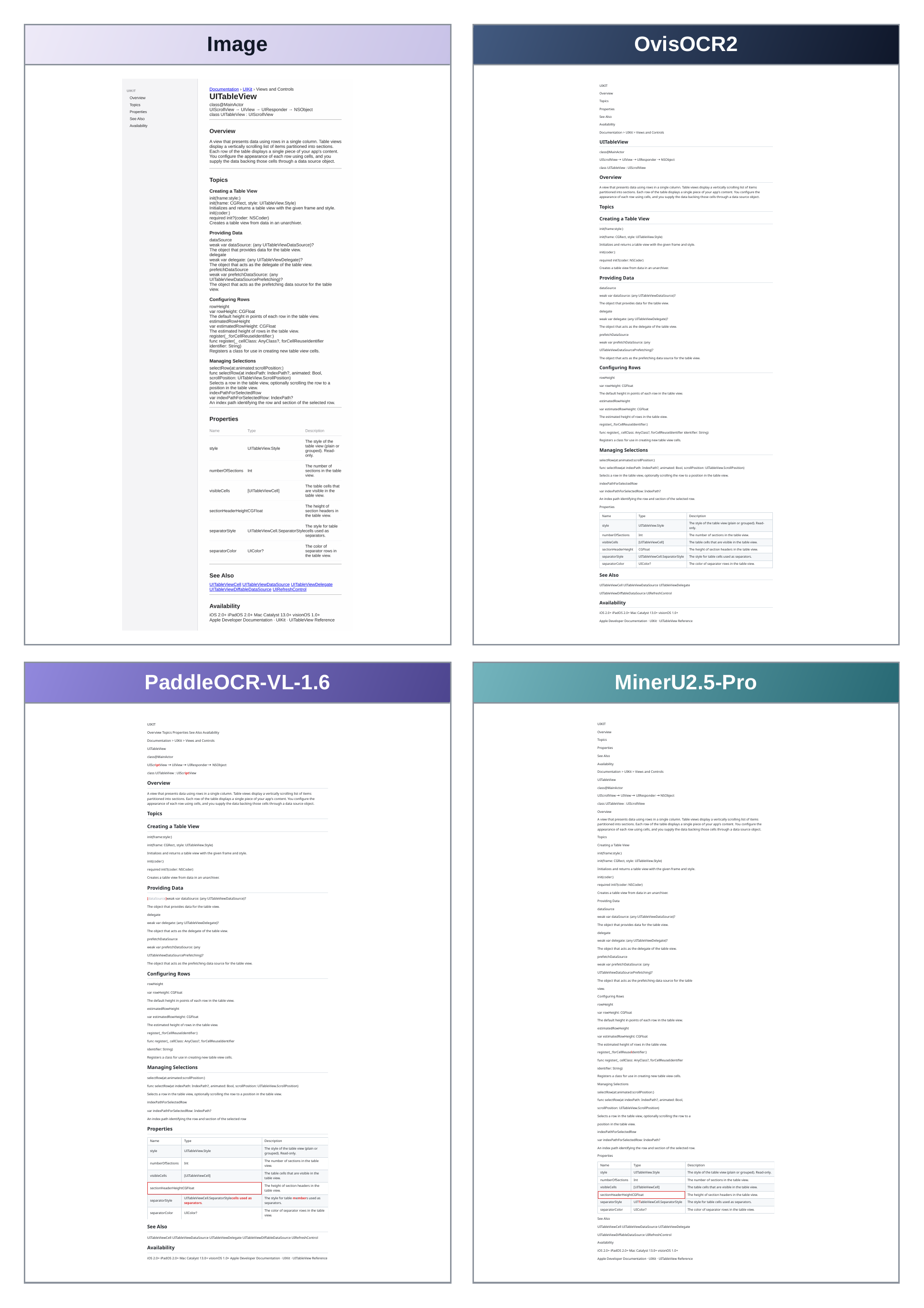}
\caption{Qualitative comparison on a table document.}
\label{fig:caset2}
\end{figure}

\clearpage
\begin{figure}[!htbp]
\ContinuedFloat
\centering
\includegraphics[width=\linewidth]{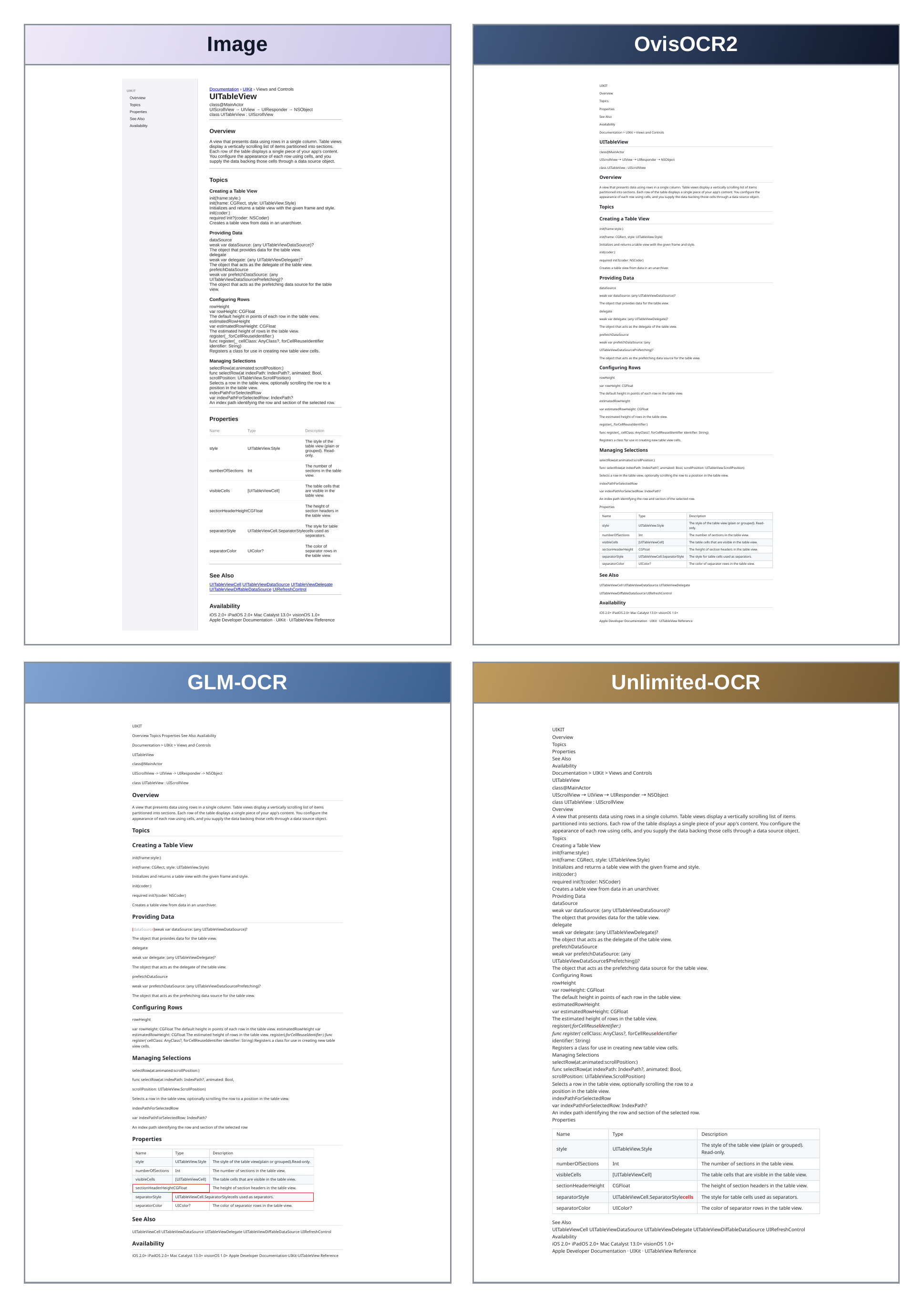}
\caption{Qualitative comparison on a table document (continued).}
\end{figure}

\clearpage
\begin{figure}[!htbp]
\centering
\includegraphics[width=\linewidth]{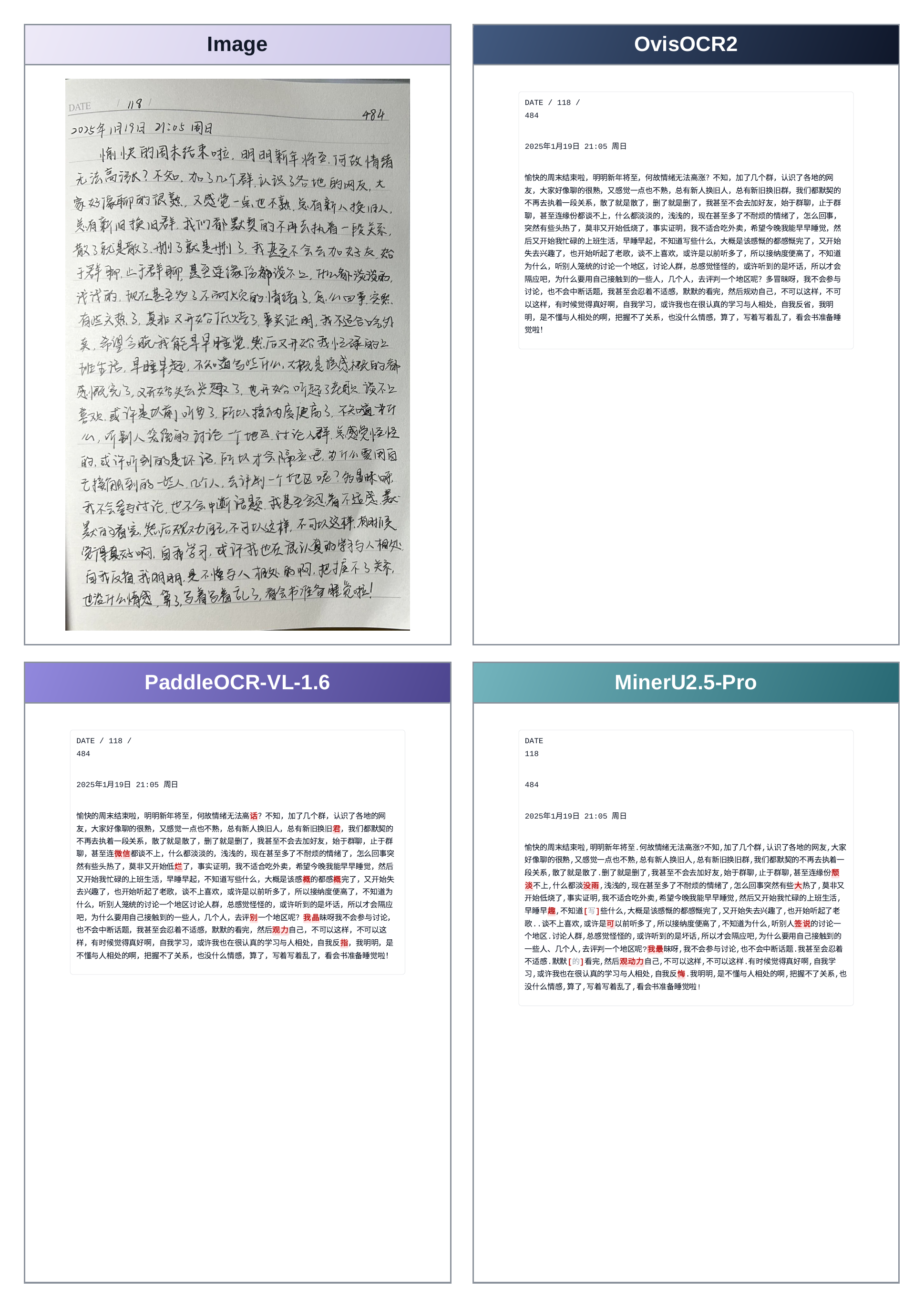}
\caption{Qualitative comparison on a handwritten document.}
\label{fig:caseh1}
\end{figure}

\clearpage
\begin{figure}[!htbp]
\ContinuedFloat
\centering
\includegraphics[width=\linewidth]{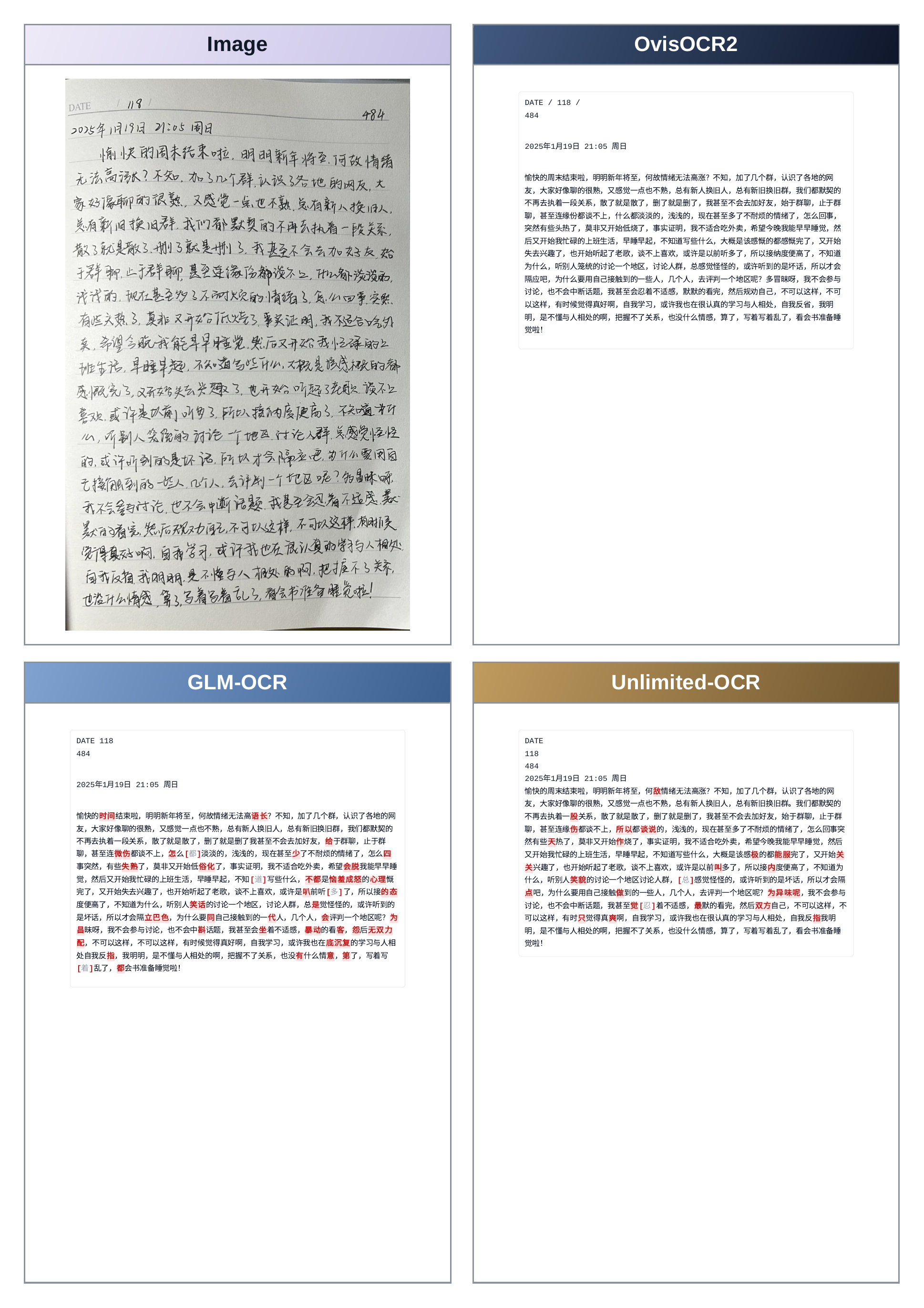}
\caption{Qualitative comparison on a handwritten document (continued).}
\end{figure}

\clearpage
\begin{figure}[!htbp]
\centering
\begin{overpic}[width=\linewidth]{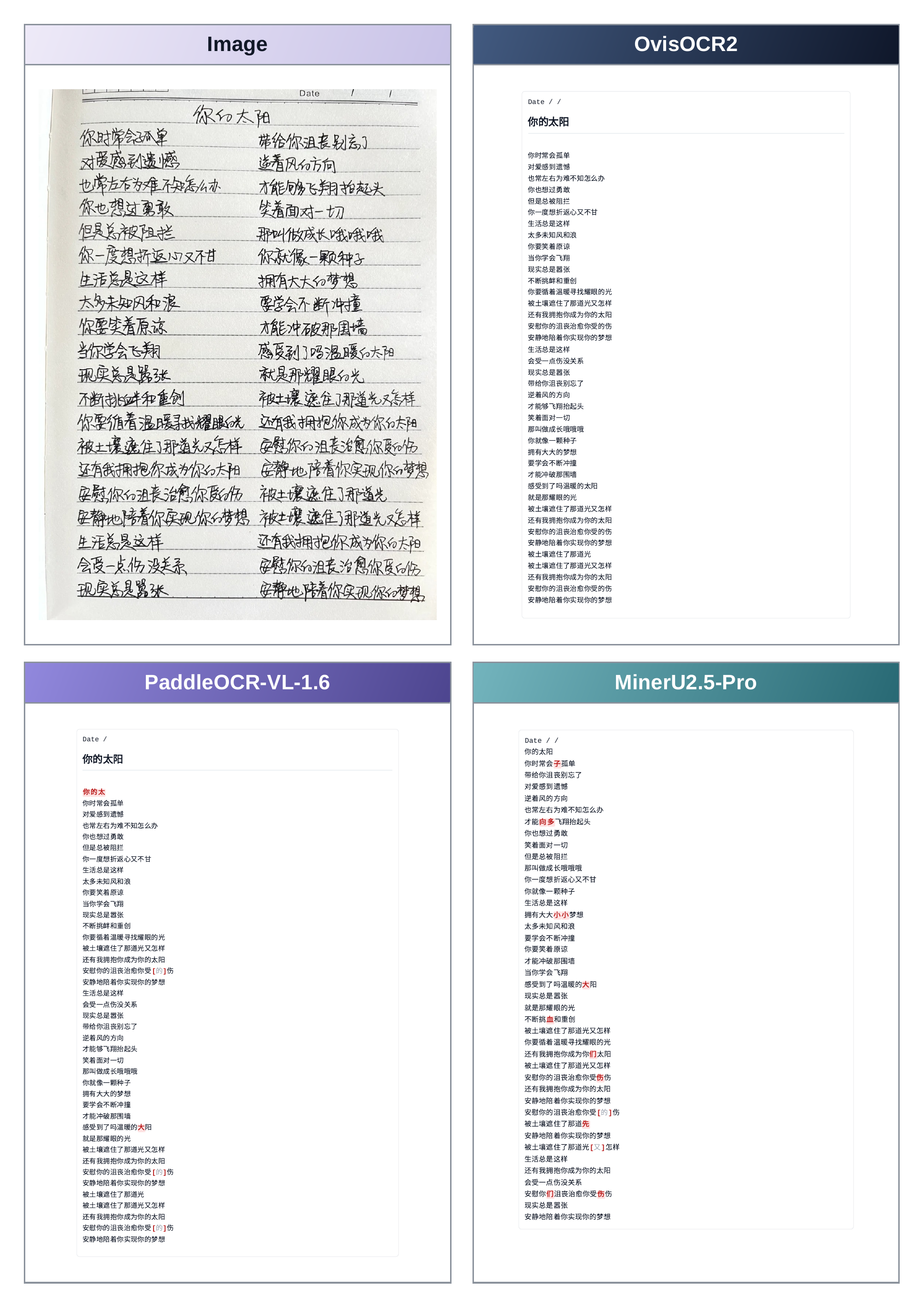}
  \put(53.3,43.0){\color[rgb]{1,0.15,0}\sffamily\fontsize{8}{9}\selectfont Reading Order Error}
\end{overpic}
\caption{Qualitative comparison on a handwritten document.}
\label{fig:caseh2}
\end{figure}

\clearpage
\begin{figure}[!htbp]
\ContinuedFloat
\centering
\begin{overpic}[width=\linewidth]{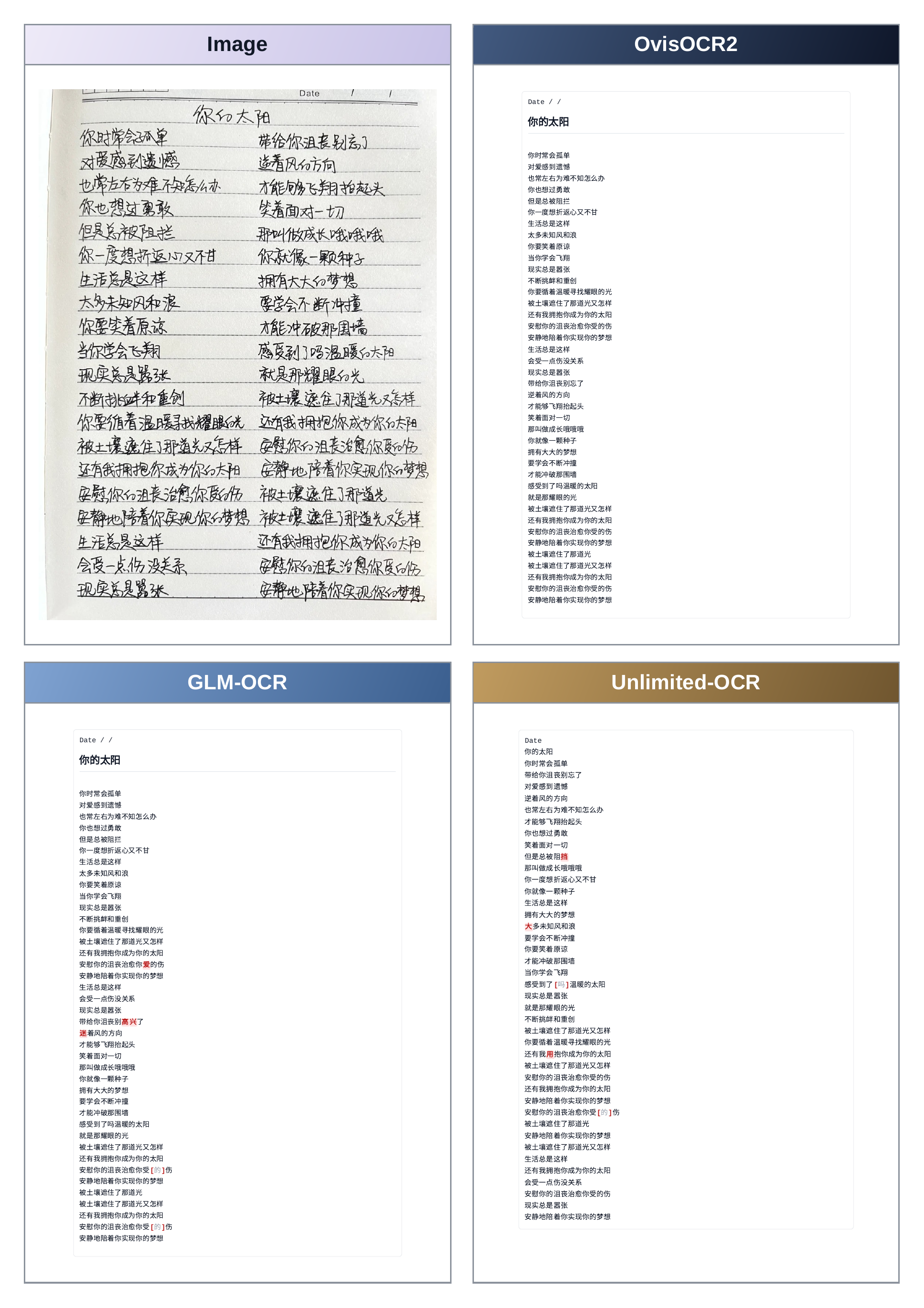}
  \put(53.3,43.0){\color[rgb]{1,0.15,0}\sffamily\fontsize{8}{9}\selectfont Reading Order Error}
\end{overpic}
\caption{Qualitative comparison on a handwritten document (continued).}
\end{figure}

\end{document}